\documentclass{article}
\usepackage{ijcai23}

\usepackage{times}
\usepackage{soul}
\usepackage{url}
\usepackage[hidelinks]{hyperref}
\usepackage[utf8]{inputenc}
\usepackage[small]{caption}
\usepackage{graphicx}
\usepackage{amsmath}
\usepackage{amsthm}
\usepackage{booktabs}
\usepackage{algorithm}
\usepackage{algorithmic}
\usepackage[switch]{lineno}

\title{Improving Sample Efficiency in Evolutionary RL Using Off-Policy Ranking}

\author{
Eshwar S R$^1$
\and
Shishir Kolathaya$^1$\and
Gugan Thoppe$^1$
\affiliations
$^1$Indian Institute of Science
\emails
\{eshwarsr, shishirk, gthoppe\}@iisc.ac.in
}

\usepackage{multirow}
\usepackage{subcaption}
\usepackage{amsmath}
\usepackage{amssymb}
\usepackage{mathtools}
\usepackage{amsthm}
\usepackage{microtype}
\usepackage{tabularx}
\usepackage[hang,flushmargin]{footmisc}
\usepackage{arydshln}

\newcommand{\bE}{\mathbb{E}}
\newcommand{\bP}{\mathbb{P}}
\newcommand{\bR}{\mathbb{R}}

\newcommand{\cA}{\mathcal{A}}

\newcommand{\cS}{\mathcal{S}}
\newcommand{\diag}{\textnormal{diag}}
\newcommand{\df}[1]{\textnormal{d}#1}

\begin{document}

\maketitle

\begin{abstract}
    Evolution Strategy (ES) is a powerful black-box optimization technique based on natural evolution. An essential step in each iteration of an ES method entails ranking candidate solutions based on some fitness score. In the context of Reinforcement Learning (RL), this ranking step requires evaluating multiple policies. This evaluation is presently done via on-policy approaches: each policy's score is estimated by interacting several times with the environment using that policy. Such approaches lead to many wasteful interactions since, once the ranking is done, only the data associated with the top-ranked policies are used for subsequent learning. To improve sample efficiency, we propose a novel off-policy alternative for ranking based on a local approximation for the fitness function. We demonstrate our idea for state-of-the-art ES methods such as Augmented Random Search (ARS) and Trust Region Evolution Strategy (TRES). Simulations in MuJoCo tasks show that, compared to the original methods, our off-policy variants have similar running times for reaching reward thresholds but need only around 70\% as much data on average. In fact, in some environments, such as HalfCheetah-v3 and Ant-v3, we need only about 50\% as much data.
\end{abstract}

\section{Introduction}
\label{introduction}
In optimization, features of the objective function such as linearity, convexity, or differentiability often are either non-existent, unknown, or impossible to detect. An Evolution Strategy (ES), due to its derivative-free nature, is a go-to alternative in such scenarios. \cite{ES} proposed the first competitive ES method in Reinforcement Learning (RL) settings. However, its effectiveness relies heavily on several complicated ideas and the usage of neural networks for parameterizing the policies. Thankfully, two recent ES methods---
the Augmented Random Search (ARS) \cite{ARS} and Trust Region Evolution Strategy (TRES) \cite{TRES}---showed that such complications may not be needed for a state-of-the-art RL approach. In particular, they demonstrated that it often suffices to 
work with only deterministic linear policies.
Our work proposes novel off-policy variants of ARS and TRES: these have running times comparable to the original ones, but need significantly less data (sometimes as less as 50\%). 

The formal motivation for our work is as follows: explore ES algorithms in RL that minimize the number of agent-environment interactions while keeping the control policy as simple as possible. Clearly, this would be of significance in practical RL. A vital example is robotics, where collecting samples is very expensive since the process involves active calibration, maintenance, and safety checks for every component, and overlooking these can result in unsafe behaviours. 


We now provide an overview of the ES philosophy. As the name suggests, ES is an iterative optimization framework inspired by evolution; see ~\cite{li2020evolution} for a recent survey. In each iteration, an ES method i.) obtains a bunch of candidate solutions from some sampling distribution, ii.) ranks them using some fitness score based on the objective function, and iii.) uses the top-ranked ones to update the sampling distribution for use in the next iteration. 

In a typical ES method for RL, a candidate solution is a specific policy, while its fitness is the associated value function. Existing techniques, including ARS, use an on-policy approach to find this fitness score: 
interact several times with the environment using the policy whose score needs to be estimated.
Since multiple policies need to be ranked in each iteration, current ES approaches end up with significantly high interactions.
Notably, most of this data is discarded in each iteration except those related to the top-ranked policies. Our proposed ARS variant improves sample efficiency by replacing this wasteful approach with an off-policy alternative. We remark that this is the first ES method in RL, where an off-policy approach is used for ranking.

Our ranking technique improves sample complexity due to two novel features: i.) the fitness function choice and ii.) the use of a kernel approximation to estimate the same. 

\textbf{Fitness Function}: Instead of using the value function $\eta(\tilde{\pi})$  of a candidate policy $\tilde{\pi}$ as its fitness score, our approach employs its approximation $L_\pi(\tilde{\pi})$ defined in terms of a different policy $\pi$ called the behavior policy ~\cite[(6)]{ATRPO}. 
Indeed,
any ranking of policies based on an approximation to the value function is going to be sub-optimal. However, it is also the key factor that enables off-policy ranking. As we shall see, the data generated by the \emph{single} policy $\pi$ can now be used to rank all the candidate policies! 
 


\textbf{Kernel Approximation}: $L_\pi(\tilde{\pi})$ is estimated in ~\cite{ATRPO} via importance sampling (see (40) in ibid). This approach works only when both $\pi$ and $\tilde{\pi}$ are stochastic. However, the sample efficiency of ARS crucially relies on the candidate policies being deterministic. Therefore, simply using stochastic policies in ARS to incorporate the importance sampling idea isn't ideal for sample efficiency. To circumvent this issue, we propose to alternatively smooth the deterministic policies using a suitable kernel function. This approach is loosely inspired from ~\cite{DeterOffKernel1,DeterOffKernel}, which studies extending ideas from discrete contextual bandit settings to the continuous case.

The role of  $L_\pi(\tilde{\pi})$ in ~\cite{ATRPO} and our work is quite different. There it is used to approximate an intermediate objective function within a policy improvement scheme, i.e., $L_\pi(\tilde{\pi})$ plays a direct role in their update rule (see (13) in ibid). Here, instead, we use it only for coarsely ranking multiple candidate policies in a sample-efficient fashion. In other words, $L_\pi(\tilde{\pi})$ needs to be accurately estimated in \cite{ATRPO}, while a rough estimate suffices  for us. Importantly, since the policies there are stochastic, a complex neural network is additionally needed for improving their sample efficiency. This complication is avoided here because our $\tilde{\pi}$'s and $\pi$ are all deterministic. 

\textbf{Key Contributions:} The main highlights of this work are 
\begin{enumerate}
    \item We propose novel off-policy variants of ARS and TRES; see Algorithms~\ref{alg:offars} and \ref{alg:offtres}. Specifically, we replace the wasteful on-policy ranking step with an efficient off-policy version. Note that this is the first usage of off-policy ranking in ES. Also, while we use this idea for ARS and TRES, it is extendable to other ES methods. 
    
    
    \item Our simulations on benchmark MuJoCo locomotion tasks (Section~\ref{s:Expt}) show that our variants reach reward thresholds in running times comparable to the original ARS and TRES. However, we often need only 50-80\% as much data. This is significant when  interactions with the environment is either hard or expensive, e.g., robotics. 

    \item We also do sensitivity to seed and hyperparameter tests similar to \cite{ARS}. Our results are similar in spirit to what was obtained in ibid. That is, the median trajectory crosses the reward threshold in most environments, confirming that our algorithm is robust to seed and hyperparameter choices. 
\end{enumerate}

\section{Relevant Background and Research Gap}
We provide here a brief overview of some of the important advances in RL relevant to our work. At the end, we also describe the current research gap that our work addresses. 

While RL has been around since the 1950s, it is only in the last decade---thanks to cheap computing and deeplearning advances---that we have seen any evidence of human and superhuman level performances. The pioneering work here was ~\cite{mnih2015human} which proposed the Deep Q Network (DQN). This algorithm combines the popular Q-learning approach with a class of artificial neural networks. In particular, this was the first work to demonstrate that instabilities due to nonlinear function approximators can be handled effectively. A major issue with DQN though is that it can only handle discrete and low-dimensional action spaces. The seminal Deep Deterministic Policy Gradient (DDPG) algorithm in ~\cite{DDPG} was proposed to overcome this issue.

In a different direction, ~\cite{TRPO} proposed a policy iteration algorithm in the discounted reward setting called Trust Region Policy Optimization (TRPO). This was recently generalized to the average reward case by \cite{ATRPO}. 
Both these methods, in each iteration, repeatedly try to identify a better policy in the neighborhood of the current estimate. The difficulty, however, is that the value functions of neighbourhood policies are unknown. \cite{TRPO} and \cite{ATRPO} resolve this  by instead finding a policy $\tilde{\pi}$ that is best with respect to a local approximation  $L_\pi(\tilde{\pi})$ of the value-function $\eta(\tilde{\pi}).$
In the Atari domain, TRPO outperforms previous approaches on some of the games. More recently, ~\cite{PPO} introduced the Proximal Policy Optimization (PPO) method. This is a descendant of TRPO, which is easier to implement and has a better sample complexity than previous approaches. 



On the ES literature side, the first competitive RL method was the one given in ~\cite{ES}. It is a derivative-free optimization method with similar or better sample efficiency than TRPO in most MuJoCo tasks. However, this advantage is a consequence of several complicated algorithmic ideas; see Section~3.4 of ~\cite{ARS} for the details. It is indeed true that this algorithm has scope for massive parallelization and, hence, can train policies faster than other methods. However, this benefit is realizable only {\em if data from multiple policies can be obtained in parallel}. This is often not the case in practical RL. For example, in robotics, one typically has access only to a single robot and, hence, can obtain data only from a single policy at any given time.


~\cite{ARS} and \cite{TRES} showed that all these complications and the usage of neural networks are often unnecessary. Instead, to obtain the state-of-the-art time and data efficiency, it suffices to work with deterministic linear policies and a basic random search with some minor tweaks. 

The LQR simulations in Section~4.4 of ~\cite{ARS}, however, showed that the current ARS implementation is not sample efficient in the optimal sense.
Our proposed ARS and TRES variants significantly improve upon sample efficiency without overly complicating the control policy.
We emphasize that, while we demonstrate our off-policy ranking idea in ARS and TRES, it should be useful in other ES methods as well.

\section{Preliminaries}
\label{s:Preliminaries}
Here, we describe our RL setup and provide an overview of the original ARS and TRES algorithms. 


\subsection{RL Setup and Important Definitions}
\label{subsec:setup}
Let $\Delta(U)$ denote the set of probability measures on a set $U.$ At the heart of RL, we have an infinite-horizon Markov Decision Process (MDP) represented by $(\cS, \cA, \bP, r, \rho_0).$ In this  tuple, $\cS$ denotes the state space, $\cA$ represents the action space,  $\bP: \cS \times \cA \to \Delta(\cS)$ and $r: \cS \times \cA \times \cS \to \bR$ are deterministic functions such that $\bP(s'|s, a) \equiv \bP(s, a, s')$ specifies the probability of moving from state $s$ to $s'$ after taking action $a$ and $r(s, a, s')$ is the one step reward obtained in this transition, and, finally, $\rho_0$ is the initial state distribution. 

The main goal in RL is to identify the `best' way to interact with the MDP. We now describe this point in some detail. Any stationary interaction strategy is defined via a policy $\pi: \cS \to \Delta(\cA)$ which gives a rule for picking an action when the environment is in a particular state. The quality of such a policy is the expected average
reward obtained on a single trajectory. Specifically, for a policy $\pi,$ i.) a trajectory refers to a sequence $\tau \equiv (s_0, a_0, s_1, a_1, \ldots)$ of states and actions, where $s_0 \sim \rho_0, a_0 \sim \pi(s_0), s_1 \sim \bP(\cdot|s_0, a_0), a_1 \sim \pi(s_1), s_2 \sim \bP(\cdot|s_1, a_1),$ and so on, and ii.) its quality is given by its value function
$    \eta(\pi) := \lim_{H \to \infty} \frac{1}{H} \bE_{\tau \sim \pi}\left[\hat{\eta}(\pi) \right]$,
where
%
    $\hat{\eta}(\pi) \equiv \hat{\eta}_H(\pi, \tau) = \sum_{t = 0}^{H-1} r(s_t,a_t, s_{t + 1})$.
%
The aforementioned goal then is to find a policy that solves 
\begin{equation}
\label{e:Opt.Problem}
    \max_\pi \eta(\pi).
\end{equation}

The sequence of states under a policy $\pi$ forms a Markov chain. If this chain has a stationary distribution $d_{\pi}$ and
\begin{equation}
    \label{e:stationary.dist}
    d_\pi(s)  = \lim\limits_{H \to \infty} \frac{1}{H}\sum_{t=0}^{H-1}\bP_{\tau \sim \pi}(s_t=s),
\end{equation} 
then $\eta(\pi) = \bE_{s \sim d_\pi, a \sim \pi}[r(s,a)].$ Note that this expression is independent of the initial distribution $\rho_0.$  


In our later discussions, we will also be using some terms related to the value function such as the state-bias function ($V_\pi$), action-bias function ($Q_\pi$), and the advantage function ($A_\pi$) given by 
\begin{align*}
    V_{\pi}(s) := {} & \bE_{\tau \sim \pi}\left[\sum_{t=0}^{\infty}(r(s_t,a_t) - \eta(\pi)) | s_0=s \right] \\
    Q_{\pi}(s,a) := {} & \bE_{\tau \sim \pi}\left[\sum_{t=0}^{\infty}(r(s_t,a_t) - \eta(\pi)) | s_0=s, a_0=a \right]
\end{align*}
and $A_{\pi}(s,a) :=  Q_{\pi}(s,a) - V_{\pi}(s),$ respectively. 

\subsection{Review of ARS}
\label{sebsec:ars_review}
In this subsection, we shall see how ARS solves the optimization problem given in \eqref{e:Opt.Problem}. 

ARS belongs to a family of iterative black-box optimization methods called random search ~\cite{RS}. Basic Random Search (BRS) is the simplest member of this family and is also where the origins of ARS lie. For ease of exposition, we first explain BRS's approach to solving \eqref{e:Opt.Problem}. Throughout this subsection, we restrict our attention to finite-horizon MDPs where the search space is some parameterized family of policies. Note that this is often the case in practical RL and is also what we deal with in our MuJoCo simulations. 

For the above setup, the problem in \eqref{e:Opt.Problem} translates to 
\begin{equation}
\label{e:Opt.ProblemARS}
    \max_{\theta} \bE_\tau [\hat{\eta}(\pi_\theta)] \equiv \bE_{\tau} [\hat{\eta}_H(\pi_\theta, \tau)]
\end{equation}
for some fixed $H.$ 
In general, this objective function need not be smooth. To circumvent this issue, BRS looks at its smooth variant and then uses the idea of a stochastic gradient ascent. Specifically, the alternative objective considered is $\bE_\delta \bE_{\tau}[\hat{\eta}(\pi_{\theta + \nu \delta})],$ where $\nu$ is a suitably fixed scalar, and $\delta$ is a random perturbation made up of i.i.d. standard Gaussian entries. Further, in each iteration, the gradient of this function  is estimated via a finite difference method. That is, $N$ random directions $\delta_1, \ldots, \delta_N$ are first generated in the parameter space, $\hat{\eta}(\pi_{\theta+\nu\delta_{k}})$ and $\hat{\eta}( \pi_{\theta-\nu\delta_{k}})$ are then estimated by interacting with the environment using $ \pi_{\theta + \nu\delta_{k}}$ and $\pi_{\theta-\nu\delta_{k}}$ for each $k,$ and finally $\frac{1}{N} \sum_{k=1}^{N}[\hat{\eta}(\pi_{\theta+\nu\delta_{k}}) - \hat{\eta}( \pi_{\theta-\nu\delta_{k}})]\delta_{k}$ is used as a proxy for the gradient at $\theta.$ BRS's update rule, thus, has the form
%
    $\theta_{j + 1} = \theta_j  + \frac{\alpha}{N} \sum_{k = 1}^N \left[\hat{\eta}(\pi_{\theta_j + \nu \delta_k}) - \hat{\eta}(\pi_{\theta_j - \nu \delta_k}) \right] \delta_k$
%
for some parameter choice $\alpha > 0.$



\cite{ARS} developed ARS by making the following changes to BRS. To begin with, they  restricted the search space to a class of deterministic and linearly parameterized policies: a policy now is 
represented by a matrix $M$ and the vector $Ms$ denotes the deterministic action to be taken at state $s$ under that policy.  Further, three modifications were made to the update rule of BRS. The first was to scale the  gradient estimate by the standard deviation of the $\hat{\eta}$ values; this yields the ARS-V1 algorithm. The second was to normalize the states, given as input to the policies, so that all state vector components are given equal importance; this yields ARS-V2. 
The final modification was to pick some $b < N$ and use only the $b$ best-performing search directions for estimating the gradient in each iteration. The first and the third step yield the ARS-V1t algorithm, while the combination of all three gives ARS-V2t. The ARS-V2t variant is of most interest to us and its update rule has the form
\begin{equation}
    \label{e:M.Update}
    M_{j + 1} = M_j +  \frac{\alpha}{b \sigma_R} \sum_{k = 1}^b [\hat{\eta}(\pi_{j, (k), +}) - \hat{\eta}(\pi_{j, (k), -})] \delta_{(k)},
\end{equation}
where $\sigma_R$ is the standard deviation of the 2b $\hat{\eta}$ values, $\delta_{(k)}$ denotes the $k$-th largest direction, decided based on the value of $\max\{\hat{\eta}(\pi_{j, k, +}), \hat{\eta}(\pi_{j, k, -})\}$ for different $k$ choices, and
%
$\pi_{j, k, +}(s) = (M_j + \nu \delta_k) \diag (\Sigma_j)^{-1/2}(s - \mu_j)$
and
$\pi_{j, k, -}(s) = (M_j - \nu \delta_k) \diag (\Sigma_j)^{-1/2}(s - \mu_j)$ with $\mu_j$ and $\Sigma_j$ being the mean and covariance of the $2bHj$ states encountered from the start of the training. 

The reason for focusing on ARS-V2t is that, in MuJoCo tasks, it typically outperforms the other ARS variants and also the previous state-of-the-art approaches such as TRPO, PPO, DDPG, the ES method from \cite{ES}, and the Natural Gradient method from ~\cite{rajeswaran2017towards}. 
%
This demonstrates that normalization of states, and then ranking and only picking the best directions for updating parameters often helps in improving the sample efficiency.
%
%
Nevertheless, in each iteration, ARS uses an on-policy technique to estimate $\hat{\eta}(\pi_{j, k, +})$ and $\hat{\eta}(\pi_{j, k, -})$ for each $k$ so that the $b$ best-performing directions can be identified. Because of this, we claim that ARS still does more interactions with the environment than what is needed. Also, in each iteration, it discards data that do not correspond to the top-ranked policies.




\begin{algorithm}[t]
  \caption{Off-policy ARS}
  
    
  \label{alg:offars}
\begin{algorithmic}[1]
  \STATE {\bfseries Setup:} State space $\bR^n,$ Action Space $\bR^p$

  \STATE {\bfseries Hyperparameters:} step-size $\alpha$, number of directions sampled per iteration $N$, standard deviation of the exploration noise $\nu$, number of top-performing directions to use $b$, bandwidth to use for kernel approximation $h$, number of behaviour policy trajectories to run $n_b$
  \STATE {\bfseries Initialize:} $M_0 = \mathbf{0} \in \bR^{p \times n}$, $\mu_0 = \mathbf{0} \in \bR^n$ and $\Sigma_0 = \mathbf{I}_n \in \bR^{n \times n}$ (identity matrix), $j=0$ 
   
  \WHILE {ending condition not satisfied}
  \STATE Sample $\delta_1,\delta_2,\dots,\delta_N$ in $\bR^{p \times n}$ with i.i.d. standard normal entries
  \STATE  Run $n_b$ number of trajectories using policy parameterized by $M_j$, resulting in $N_d$ number of interactions
  \STATE  Sort the directions $\delta_k$ based on $f_{\pi_{j}}(\delta_k, h)$ scores (using \eqref{eq:fpifromdeltak}), denote by $\delta_{(k)}$ the $k$-th largest direction, and by $\pi_{j,(k),+}$ and $\pi_{j,(k),-}$ the corresponding policies
  \STATE Collect $2b$ rollouts of horizon H and their corresponding return ($\hat{\eta}(\cdot)$) using the $2b$ policies 
  $$\pi_{j,(k),+}(s) = (M_j+\nu\delta_{(k)}) \diag(\Sigma_j)^{-1/2}(s-\mu_j)$$
  $$\pi_{j,(k),-}(s) = (M_j-\nu\delta_{(k)}) \diag(\Sigma_j)^{-1/2}(s-\mu_j)$$\\
  \STATE Make the update step: $$M_{j+1} = M_j + \frac{\alpha}{b \sigma_R}\sum_{k=1}^b[\hat{\eta}(\pi_{j,(k),+}) - \hat{\eta}(\pi_{j,(k),-})]\delta_{(k)},$$ where $\sigma_R$ is the standard deviation of $2b$ returns used in the update step.\\
  \STATE Set $\mu_{j+1}$, $\Sigma_{j+1}$ to be the mean and covariance of the $2bH(j+1)$ states encountered from the starting of training\\
  \STATE $j \leftarrow j+1$ \\
  \ENDWHILE
   
\end{algorithmic}
\end{algorithm}

\subsection{Overview of TRES}
\label{sebsec:tres_overview}
TRES \cite{TRES} is another state-of-the-art ES method that has been shown to outperform methods such as TRPO, PPO, and the ES method from \cite{ES}. While broadly similar, the major differences between TRES and ARS is in the choice of the objective function. Instead of using a simple sum of rewards on a single trajectory as in \eqref{e:Opt.ProblemARS}, TRES uses a novel local approximation (\cite[(25)]{TRES}) to the value function that matches the latter up to the first-order.
 The main advantage of this alternative choice is that it guarantees monotonic improvement in successive policy estimates. Also, this new objective function enables TRES to use the same data from top performing directions to update the parameters multiple times in the same iteration. This is not possible with the objective function used in ARS, and is claimed to be core reason for improving sample efficiency. A detailed review of TRES is given in Appendix \ref{tres_review}.


\section{Off-policy ARS and TRES}
\label{s:Method}
Here, we provide a detailed description of our proposed approach to improve upon the wasteful on-policy ranking step in ARS. We use the same idea also to improve upon TRES, but leave these details to Appendix \ref{op_tres}. 

Intuitively, in each iteration of our ARS variant, we plan to identify a suitable deterministic policy, interact with the environment using just this {\em single} policy, and then use the resultant data to rank the $2N$ deterministic policies $\{\pi_{j, k, +}, \pi_{j, k, -}: 1 \leq k \leq N\}.$ As a first step, we come up with a way to approximate the value function of a deterministic policy $\tilde{\pi}$ in terms of another deterministic policy $\pi.$ We focus on deterministic policies here since ARS's performance crucially depends on this determinism.

If $\pi$ and $\tilde{\pi}$ were stochastic in nature, then such an approximation has already been given in \cite{ATRPO}, which itself is inspired from similar estimates given in \cite{Kakade02approximatelyoptimal} and \cite{TRPO}. We now discuss the derivation of this approximation. 

Consider the average reward RL setup described in Section~\ref{subsec:setup}. Suppose that, for every stationary policy $\pi,$ the induced Markov chain is irreducible and aperiodic and, hence, has a stationary distribution $d_\pi$ satisfying \eqref{e:stationary.dist}. In this framework, \cite[Lemma~1]{ATRPO} showed that the value functions of the two stochastic policies $\pi$ and $\tilde{\pi}$ satisfy
\begin{equation}
    \label{eq:kakade_identity}
    \eta(\tilde{\pi}) = \eta(\pi) + \bE_{s \sim d_{\tilde{\pi}}, a \sim \tilde{\pi}}\left[A_{\pi}(s,a)\right]. 
\end{equation}
Given this relation, a natural question to ask is whether $\eta(\tilde{\pi})$ can be estimated using just the data collected by interacting with the environment using $\pi.$ The answer is no, mainly because the expectation on the RHS is with respect to the states being drawn from $d_{\tilde{\pi}}.$ In general, this distribution is not known a priori and is also hard to estimate unless you interact with the environment with $\tilde{\pi}$ itself.

Inspired by  \cite{Kakade02approximatelyoptimal} and \cite{TRPO}, \cite{ATRPO} proposed using 
\begin{equation}
\label{e:local.Approximation}
    L_\pi(\tilde{\pi}) := \eta(\pi) + \bE_{s \sim d_{\pi}, a \sim \tilde{\pi}} \left[A_{\pi}(s,a)\right]
\end{equation}
as a proxy for the RHS in \eqref{eq:kakade_identity} to overcome the above issue. There are two main reasons why this was a brilliant idea. The first is that $L_\pi(\tilde{\pi}),$ since it uses $d_\pi$ instead of $d_{\tilde{\pi}},$ can be estimated from only environmental interactions involving $\pi.$ Second, and importantly, \cite[Lemmas~2, 3]{ATRPO} showed that $|L_\pi(\tilde{\pi}) - \eta(\tilde{\pi})|$ is bounded by the total variation distance between $\pi$ and $\tilde{\pi}.$ Thus, when $\pi$ and $\tilde{\pi}$ are sufficiently close, an estimate for $L_{\pi}(\tilde{\pi})$ is also one for $\eta(\tilde{\pi}).$ Hence, $L_\pi(\tilde{\pi})$ paves the way for estimating $\eta(\tilde{\pi})$ in an off-policy fashion, i.e., using data from a different policy $\pi$. Henceforth, we refer to the policy chosen for interaction (e.g., $\pi$ above) as the behavior policy and the one whose value needs to be estimated (e.g., $\tilde{\pi}$ above) as the target policy.

We now extend the above idea to the case with deterministic policies, which we emphasize is one of our main contributions. While the idea may look simple on paper, the actual extension is not at all straightforward. The key issue is that, in the case of stochastic policies, the idea of importance sampling and, in particular, the relation
%
    $\bE_{s \sim d_{\pi}, a \sim \tilde{\pi}} \left[A_{\pi}(s,a)\right] = \bE_{s \sim d_{\pi}, a \sim \pi} \left[\frac{\tilde{\pi}(a|s)}{\pi(a|s)}A_{\pi}(s,a)\right]$
%
is used for estimating the second term in \eqref{e:local.Approximation}. However, for deterministic policies, the ratio $\tilde{\pi}(a|s)/\pi(a|s)$ will typically be $0,$ which means the estimate for the second term will also almost always be zero. Hence, this importance sampling idea for estimating $L_\pi(\tilde{\pi})$ fails for deterministic policies. 


The alternative we propose is to modify the definition of $L_{\pi}(\tilde{\pi})$ so that it becomes useful even for deterministic policies. Specifically, we redefine $L_\pi(\tilde{\pi})$ as 
\begin{equation}
\label{e:L.new.defn}
    L_\pi(\tilde{\pi}) = \eta(\pi) + \textstyle{ \mathbb{E}_{s \sim d_\pi, a \sim\pi} \left[ \frac{K_h(\|a-\tilde{\pi}(s)\|)}{K_h(\|a-\pi(s)\|)} A_\pi(s,a) \right]},
\end{equation}
where $K_h(u) = h^{-1}K(u/h)$ and $K: \bR \to \bR$ denotes a suitably chosen kernel function satisfying $\int K(u) \df{u} = 1$ and $\int u K(u) \df{u} = 0.$ This approach is loosely inspired from \cite{DeterOffKernel1,DeterOffKernel} which look at extending policy evaluation and control algorithms from discrete contextual bandit settings to the continuous case. 

While there are multiple choices for $K,$ we use $K(u) = e^{-u^2}$ in this work. Substituting this definition in \eqref{e:L.new.defn} gives
    $L_\pi(\tilde{\pi}) = \eta(\pi) + \bE_{s \sim d_{\pi}} \left[e^{-\|\pi(s) - \tilde{\pi}(s)\|^2/h^2} A_\pi(s, \pi(s))\right]$.

\begin{table*}[ht]
\centering
\begin{tabular}{|r@{\hspace{1mm}}:r@{\hspace{1mm}}|c@{\hspace{0.5mm}}:c@{\hspace{1mm}}c@{\hspace{1mm}}c@{\hspace{2mm}}|c@{\hspace{1mm}}c@{\hspace{1mm}}c@{\hspace{2mm}}|c@{\hspace{1mm}}:c@{\hspace{1mm}}c@{\hspace{1mm}}c@{\hspace{2mm}}|c@{\hspace{1mm}}c@{\hspace{1mm}}c|}
\hline
\multirow{2}{*}{Env.(-v3)}& \multirow{2}{*}{Th.} & \multicolumn{4}{c|}{ARS} & \multicolumn{3}{c|}{OP-ARS} & \multicolumn{4}{c|}{TRES} & \multicolumn{3}{c|}{OP-TRES}\\
\cline{3-16}& & N & b & Intx. & R[8] & Intx. & R[8] & \% & N & b & Intx. & R[8] & Intx. & R[8] & \% \\
\hline
Swimmer & 325 & 2&1 &520 (580) & 5 (5) & \textbf{440} & 5 & 85 & 4&2 &800 (560) & 8 (7) & 800 & 7 & 100\\
Hopper & 3120 & 8&4 & 883 (1098) & 6 (5) & \textbf{765} & 6 & 86 & - & - & - & - & - & - & -\\
HalfCheetah & 3430 & 32&4 & 4480 (3840) & 8 (7)  & 2400 & 7 & 53 & 16&8 & 2720 (2400) & 8 (8) & \textbf{1275} & 8 & 46 \\
$^\#$ Walker2d & 4390 & 40&30 & 18802 (23151) & 4 (5) & \textbf{14414} & 4 & 76 & - & - & - & - & -& - & -\\
Ant & 3580 & 60&20 & 10492 (17711) & 8 (2) & 15071* & 8 & 143 & 40&20 &10849 (10409) & 8 (8) & \textbf{6165} & 7 & 56\\
Humanoid & 6000 & 350&230 & 40852 (23594) & 5 (6) & \textbf{14260} & 6 & 35 & - & - & - & - & - & - & -\\
\hline\end{tabular}

\caption{\label{tab:table_offars_offtres} Comparison of median number of interactions for ARS, TRES, OP-ARS and OP-TRES on MuJoCo locomotion tasks to achieve prescribed reward thresholds(Th.).
R column represents the number of seeds in which the threshold was reached. \% column represents the percentage of data required by our method compared to the original method. The * in Ant-v3 signifies the interleaving of on-policy evaluations. The \# signifies 16 seeds were used instead of 8 in Walker2d environment for ARS.}
\end{table*}

\begin{figure*}[ht]
     \centering
     \begin{subfigure}[c]{0.27\textwidth}
         \centering
         \includegraphics[width=\textwidth]{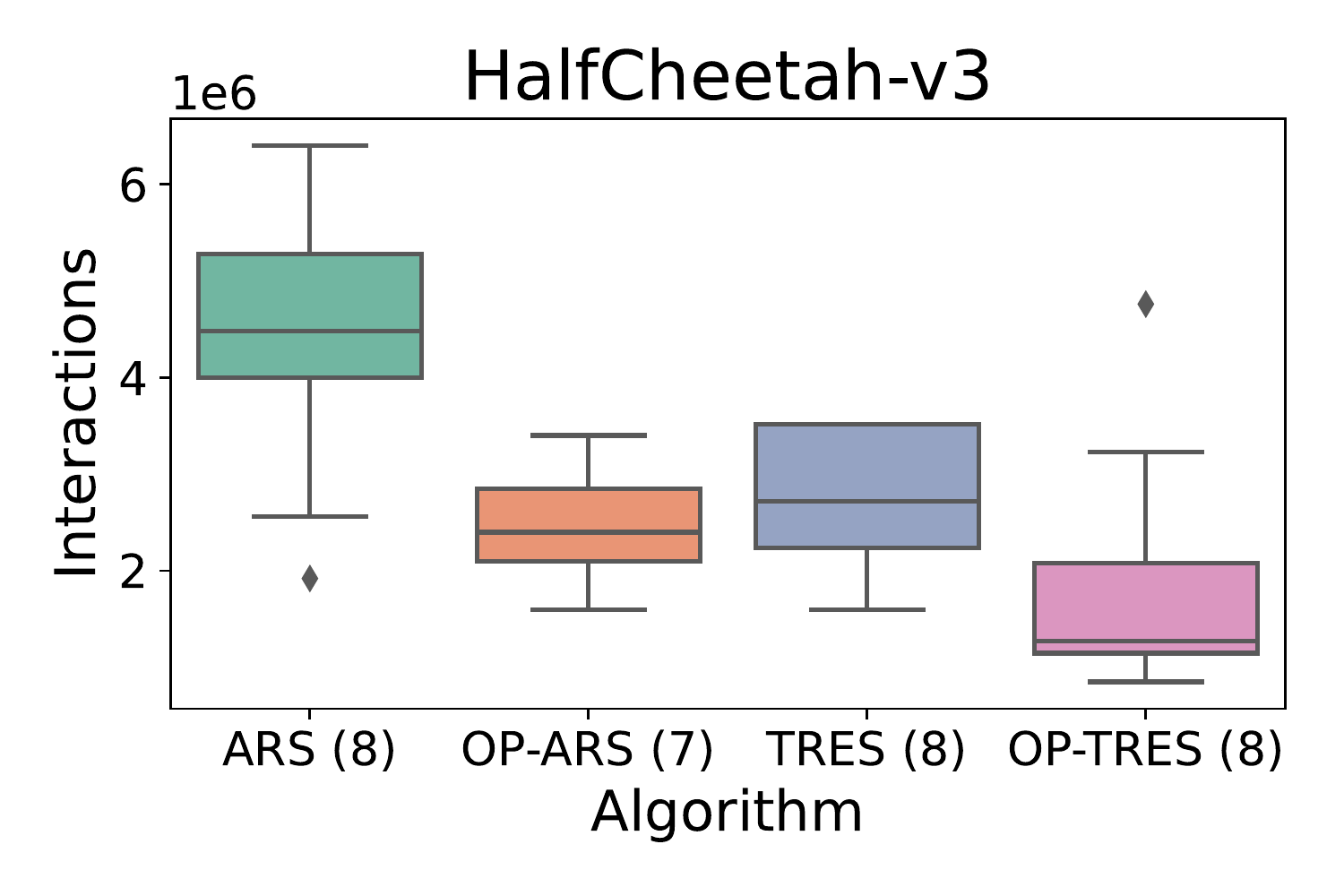}
     \end{subfigure}
     \begin{subfigure}[c]{0.27\textwidth}
         \centering
         \includegraphics[width=\textwidth]{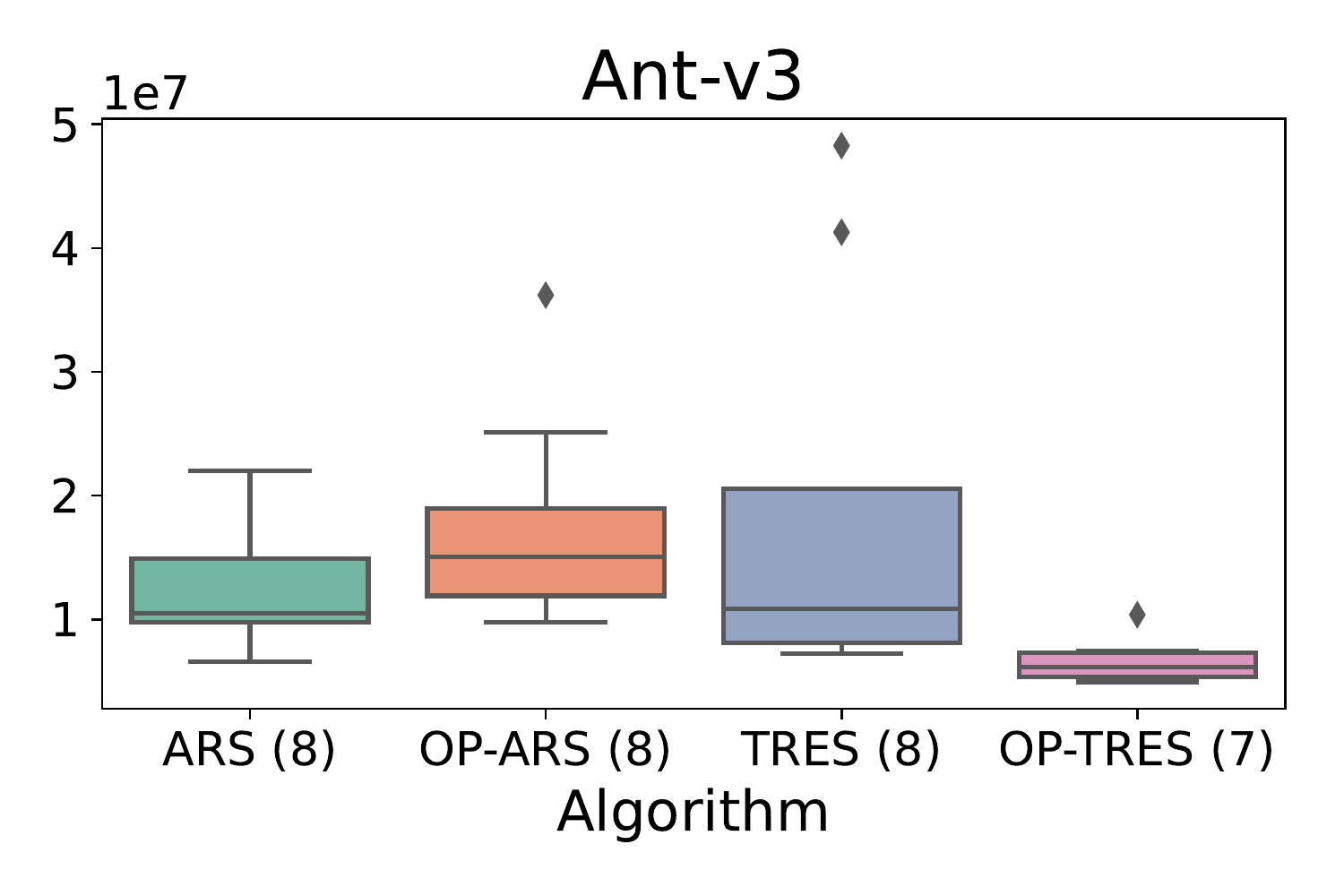}
     \end{subfigure}
     \begin{subfigure}[c]{0.27\textwidth}
         \centering
         \includegraphics[width=\textwidth]{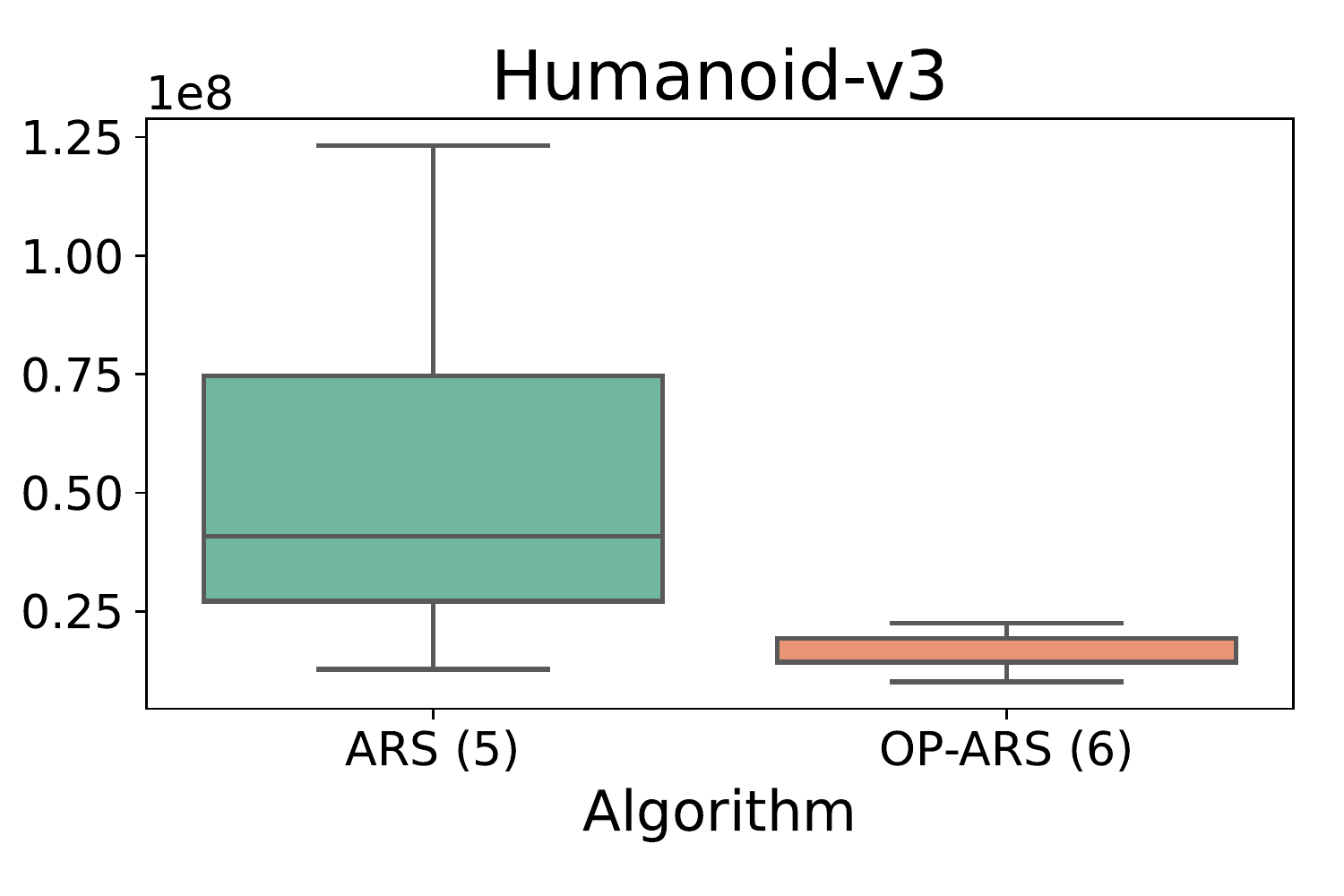}
     \end{subfigure}
        \caption{Box plots of number of interactions required to reach the reward threshold in ARS, TRES, OP-ARS and OP-TRES. The number next to the algorithm's name represents the number of seeds in which the threshold was reached.}
        \label{fig:box_plots}
\end{figure*}

The reason for the above choice of $K$ is that it performs quite well in simulations and, importantly, provides a clean intuitive explanation for $L_\pi(\tilde{\pi}).$ That is, $L_\pi(\tilde{\pi})$ assigns a higher value to a policy $\tilde{\pi}$ if it takes actions similar to $\pi$ at all those states $s$ where $A_\pi(s, \pi(s))$ is large. In summary,   $L_\pi(\tilde{\pi})$ given above provides us with the desired expression to approximate the value function of a deterministic target policy with only the data from a deterministic behavior policy. 


We now discuss incorporating this expression in ARS to improve its sample efficiency. In particular, we now show how we can rank the $2N$ policies $\{\pi_{j, k, +}, \pi_{j, k, -}\},$ generated randomly in each iteration of ARS, in an off-policy fashion (see Section~\ref{sebsec:ars_review} for further details on ARS). 

The first thing we need to decide is the choice of the behavior policy for interacting with the environment. Recall from the discussion below \eqref{e:local.Approximation} that $|\eta(\tilde{\pi}) - L_\pi(\tilde{\pi})|$ is small when $\pi$ and $\tilde{\pi}$ are sufficiently close. Now, since the policy parameterized by $M_j$ is close to each of the $2N$ policies specified above, it is the natural choice for the behavior policy and, indeed, this is what we use.

In summary, our ranking in each iteration works as follows:
\begin{enumerate}
    \item Interact with the environment using the behavior policy $\pi_j \equiv \pi_{M_j}$ on $n_b$ number of trajectories. Each trajectory here is presumed to have $H$ many time steps (or less in case of premature termination). Suppose these interactions result in $N_d$ $(s_t, a_t, r_t, s_{t + 1})$ transitions overall.
    
    \item Estimate $Q_{\pi_j}(s_t, a_t),$ $0 \leq t \leq N_d - 1,$ using the definition given in Section~\ref{subsec:setup}. 
    
    \item Estimate 
    \begin{align}
         f_{\pi_{j}}(\delta_k, h) = {} & \mathbb{E}[e^{-\|\nu \delta_k s\|^2/h^2} Q_{\pi_{j}}(s,a)] \nonumber
        \\
         \approx {} & \frac{1}{N_d} \sum_{t=0}^{N_d-1}[ e^{-\|\nu \delta_k s\|^2/h^2} Q_{\pi_{j}}(s_t,a_t)] \label{eq:fpifromdeltak}
    \end{align}
    for each $1 \leq k \leq N.$ Note that $f_{\pi_j}(\delta_k, h)$ is a proxy for the expression in \eqref{e:L.new.defn}. In that, it ignores all the constant terms: those that depend only on the behavior policy $\pi_j.$ 
    
    \item Use the above estimates to rank $\{\pi_{j, k, +}, \pi_{j, k, -}\}.$
\end{enumerate}

Once the $b$ best--performing directions are identified, the rest of our ARS variant more or less proceeds as the original. That is, we come up with  better estimates of the value-functions of these top policies in an on-policy fashion and improve upon $M_j$ along the lines discussed in \eqref{e:M.Update}. The complete details are given in Algorithm~\ref{alg:offars}. A detailed section discussing the differences in the original ARS from our off-policy variant is given in Appendix \ref{appendix:ars_vs_ours}.

We end our discussion here by pointing out that the original ARS used $2N$ interaction trajectories, each of length roughly $H,$ in each iteration. In our variant, we only need $2b + n_b$ many trajectories. When $b < N,$ this difference is what leads to the significant reduction in interactions seen in our simulations.


\section{Experiments}
\label{s:Expt}

\begin{figure*}[h]
     \centering
     \begin{subfigure}[c]{0.27\textwidth}
         \centering
         \includegraphics[width=\textwidth]{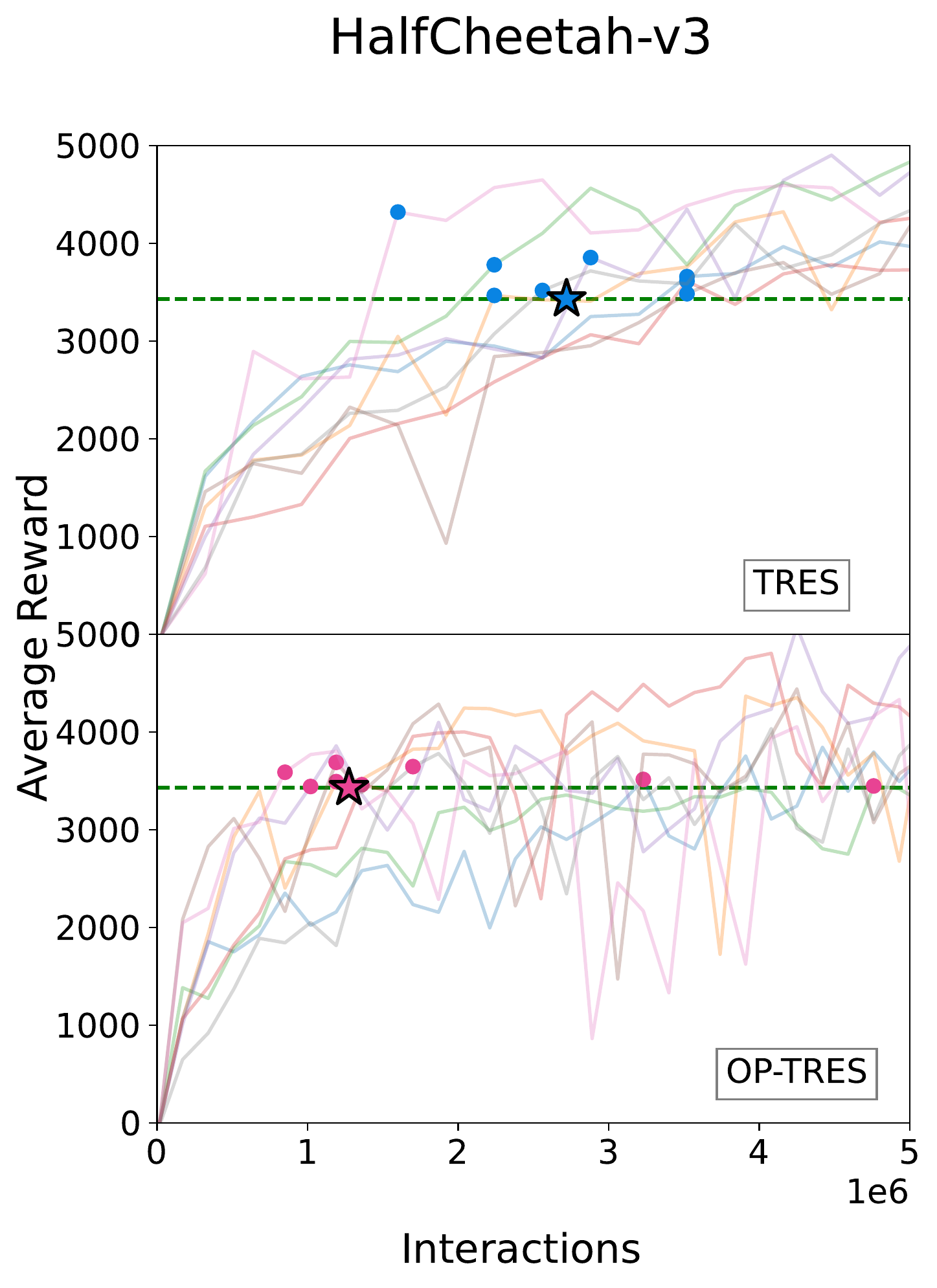}
     \end{subfigure}
     \begin{subfigure}[c]{0.265\textwidth}
         \centering
         \includegraphics[width=\textwidth]{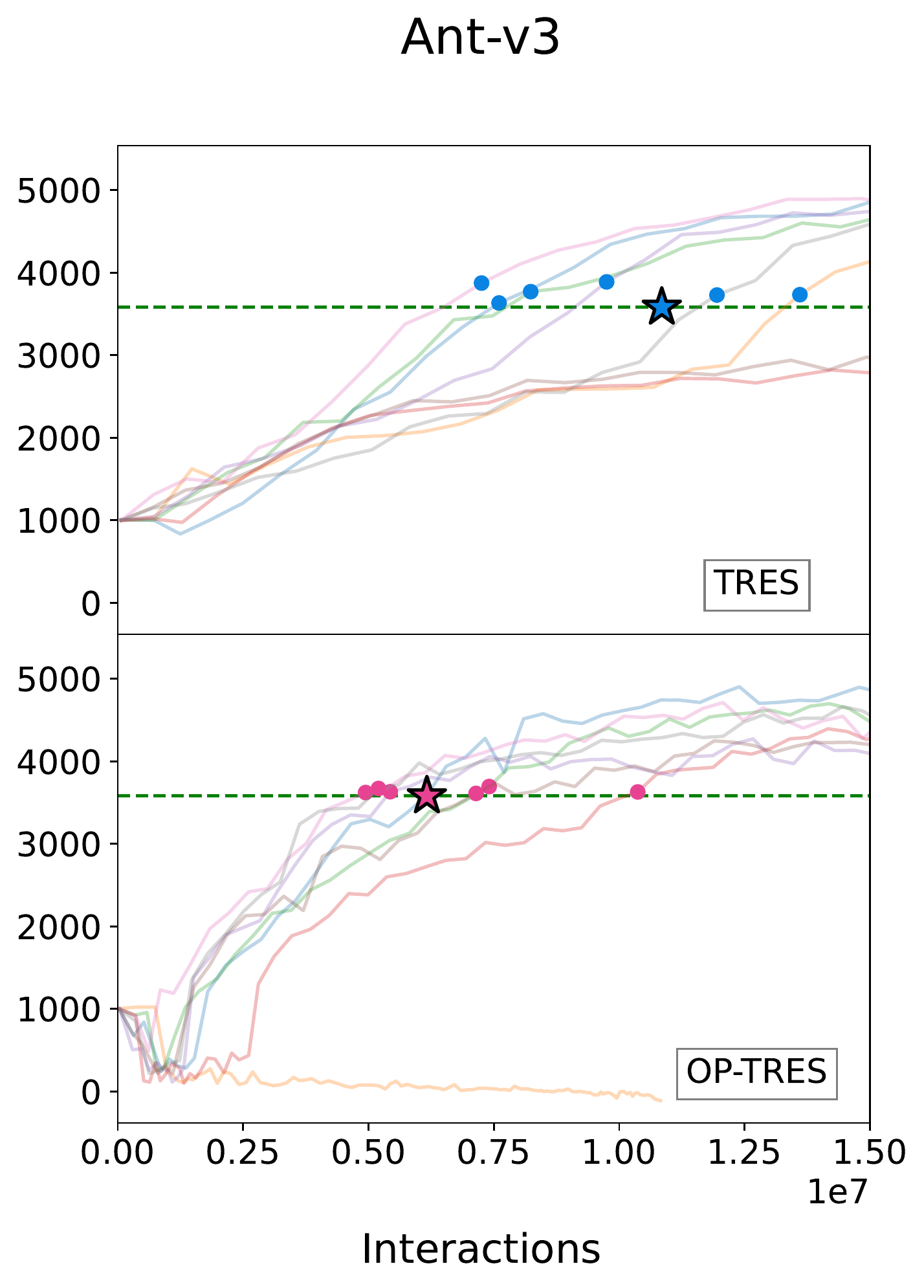}
     \end{subfigure}
     \begin{subfigure}[c]{0.26\textwidth}
         \centering
         \includegraphics[width=\textwidth]{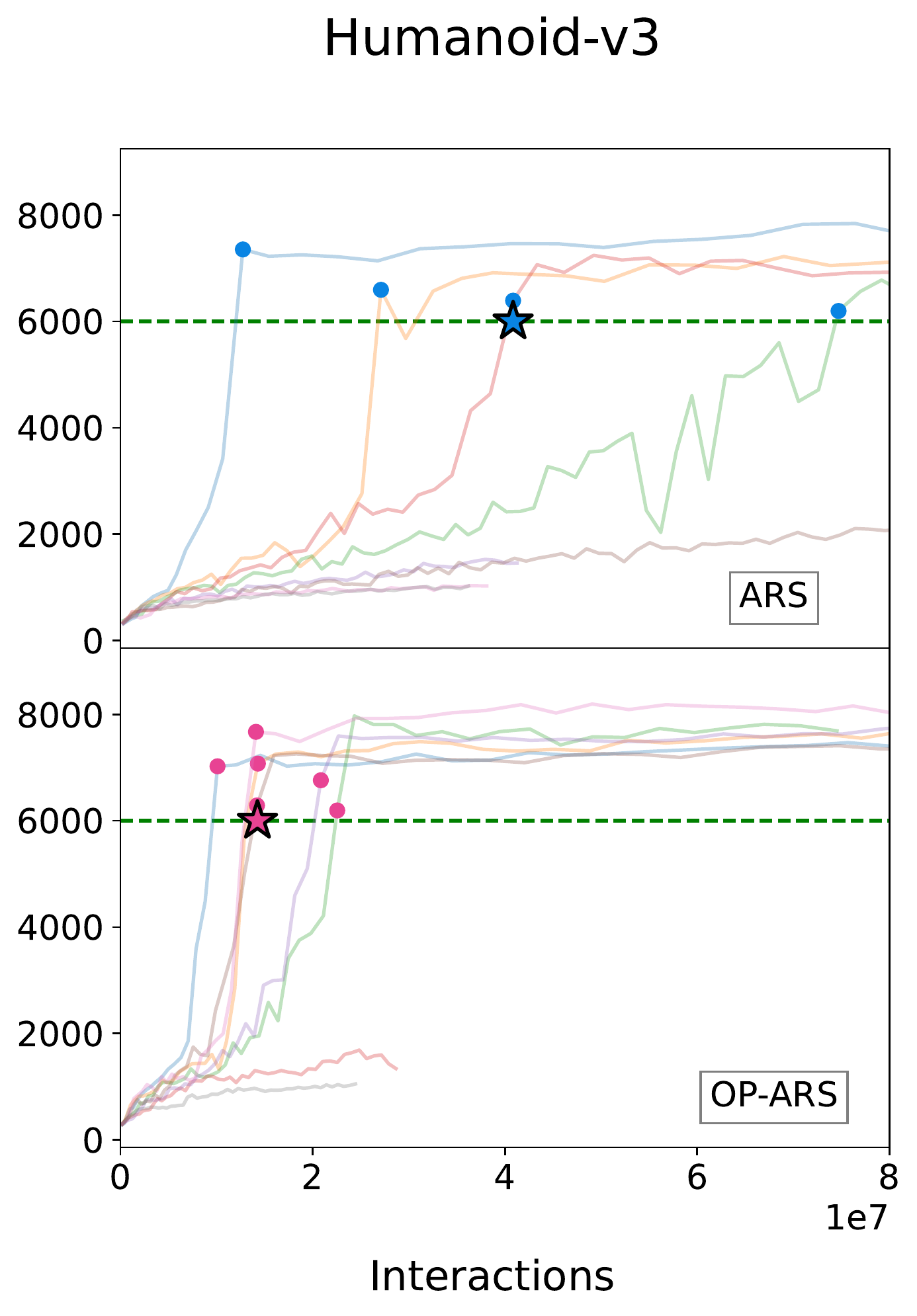}
     \end{subfigure}
        \caption{Figures representing the trajectories of various runs of ARS, OP-ARS, TRES and OP-TRES algorithms where the number of interactions with the environment is plotted against the average reward. 
        }
        \label{fig:all_seeds}
\end{figure*}

\begin{figure*}[h!]
     \centering
     \begin{subfigure}[c]{0.27\textwidth}
         \centering
         \includegraphics[width=\textwidth]{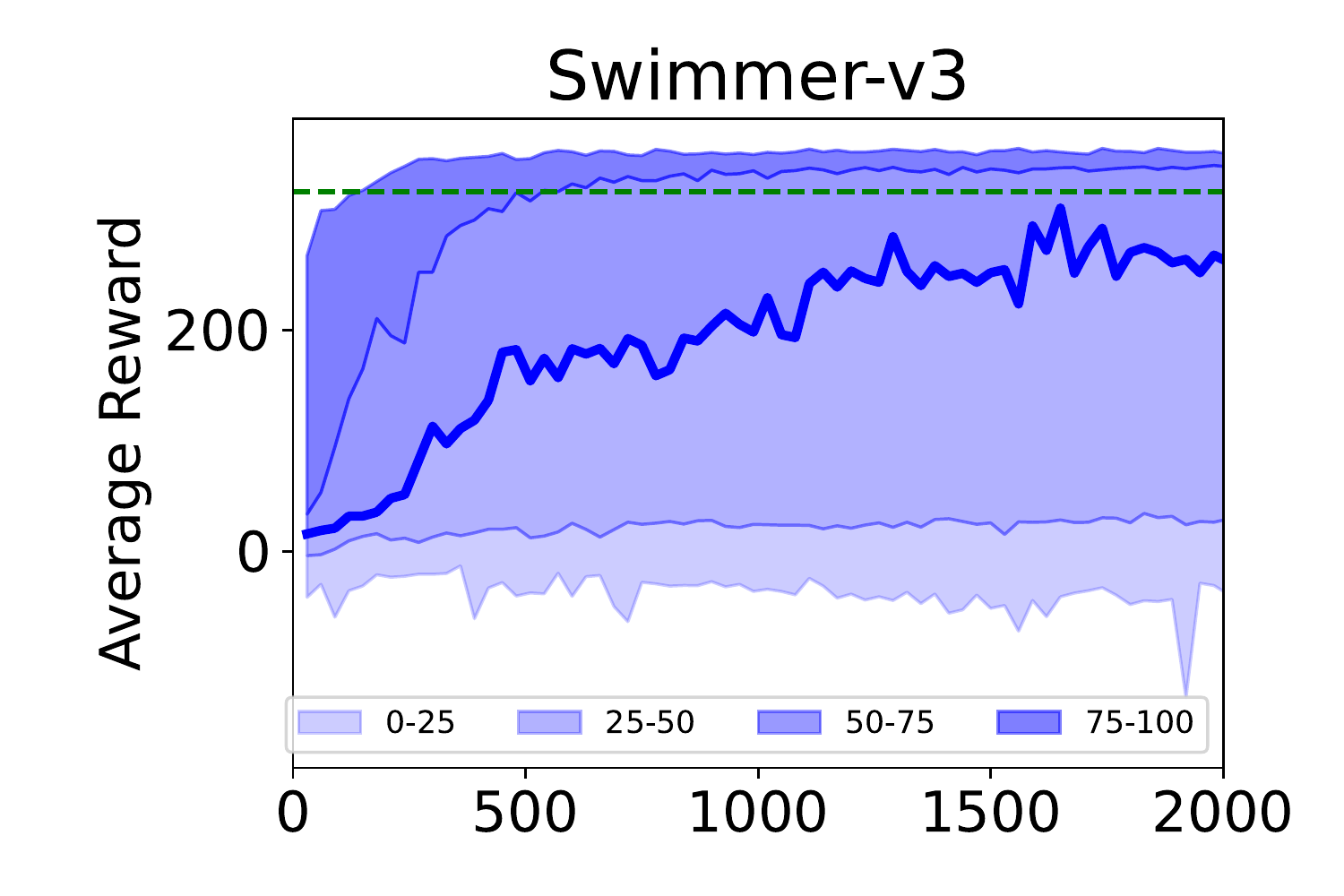}
         \caption*{}
     \end{subfigure}
     \begin{subfigure}[c]{0.27\textwidth}
         \centering
         \includegraphics[width=\textwidth]{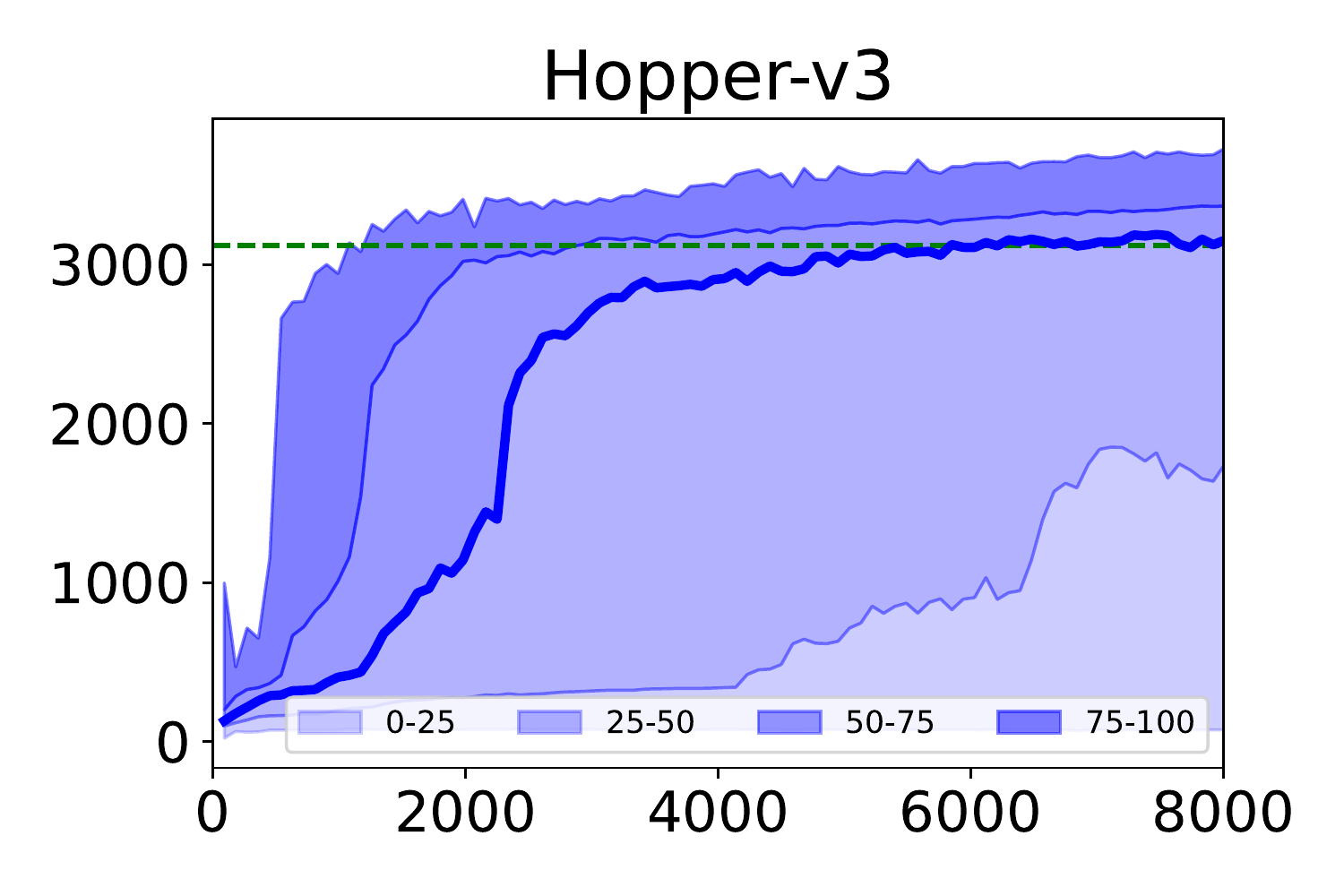}
         \caption*{(a) 100 seeds experiment}
     \end{subfigure}
     \begin{subfigure}[c]{0.27\textwidth}
         \centering
         \includegraphics[width=\textwidth]{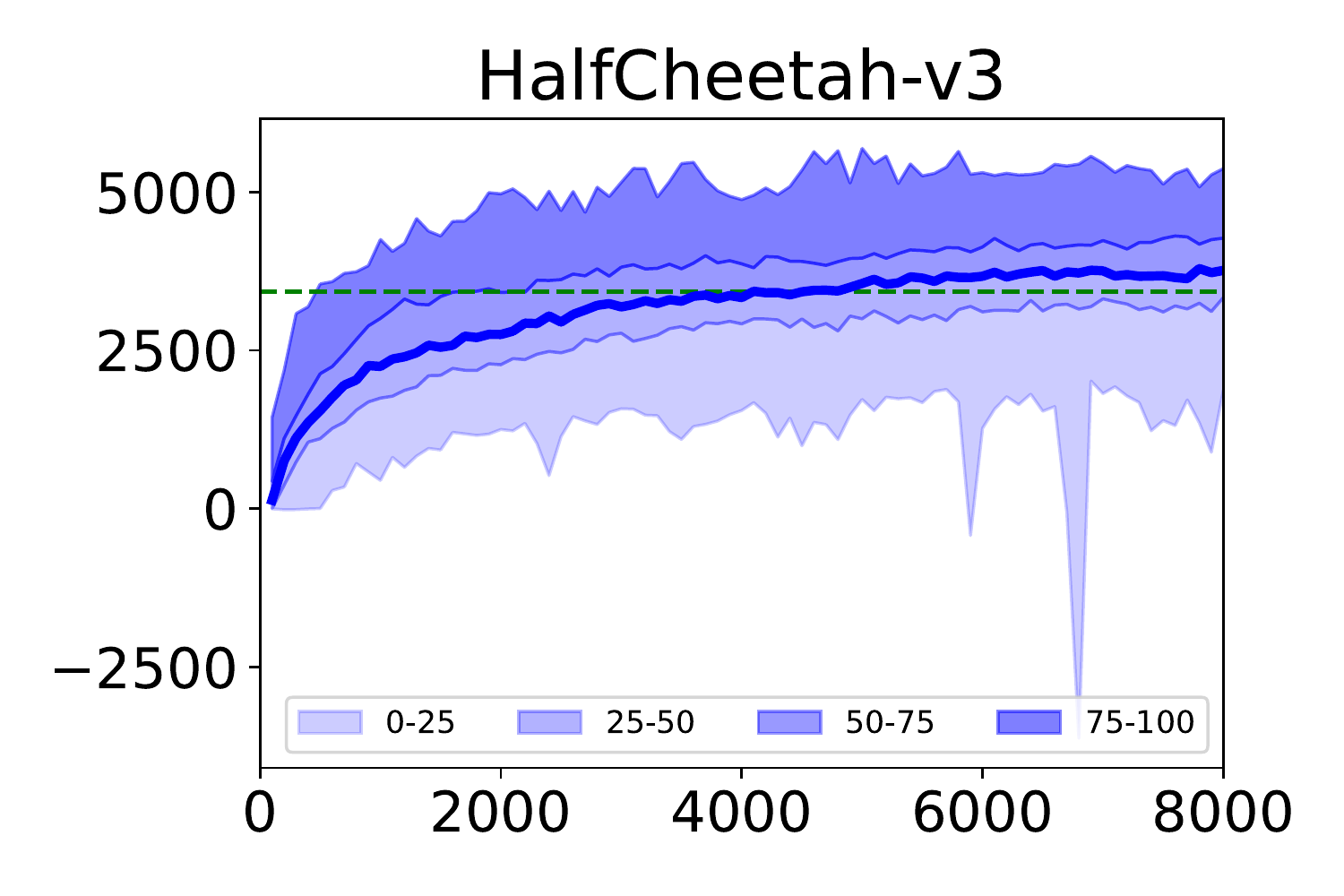}
         \caption*{}
     \end{subfigure}
     
     \begin{subfigure}[c]{0.27\textwidth}
         \centering
         \includegraphics[width=\textwidth]{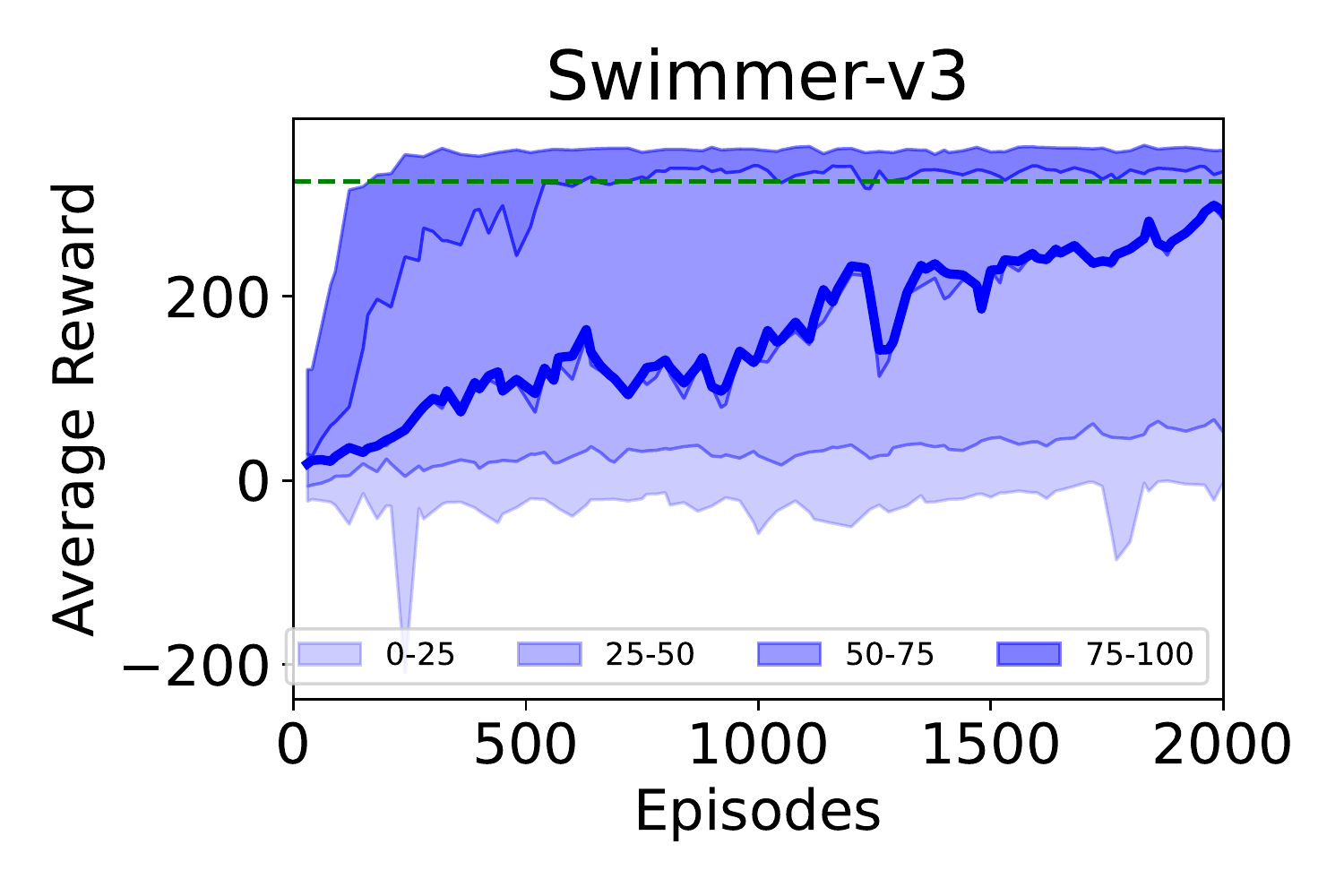}
         \caption*{}
     \end{subfigure}
     \begin{subfigure}[c]{0.27\textwidth}
         \centering
         \includegraphics[width=\textwidth]{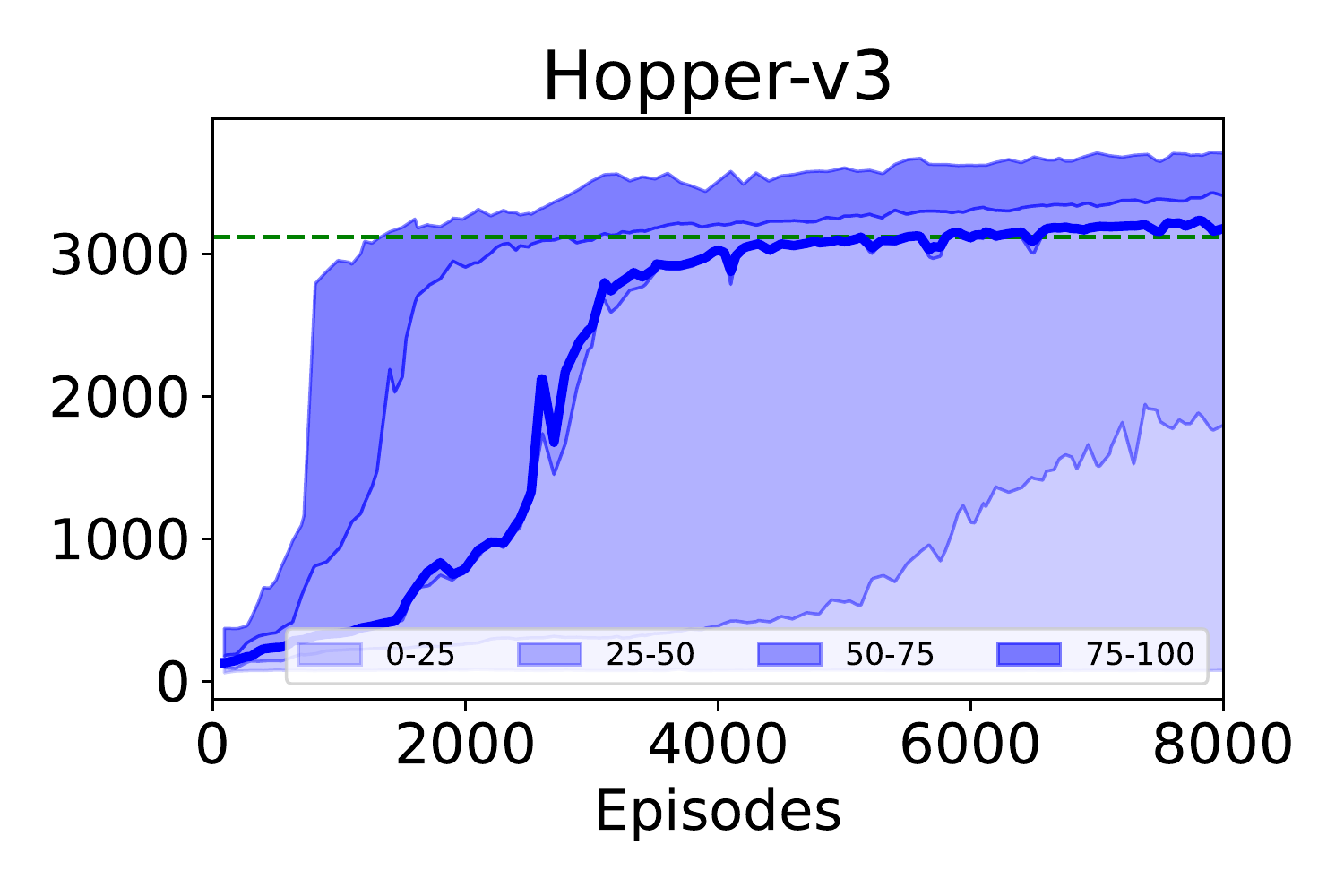}
         \caption*{(b) Hyperparameter sensitivity test}
     \end{subfigure}
     \begin{subfigure}[c]{0.27\textwidth}
         \centering
         \includegraphics[width=\textwidth]{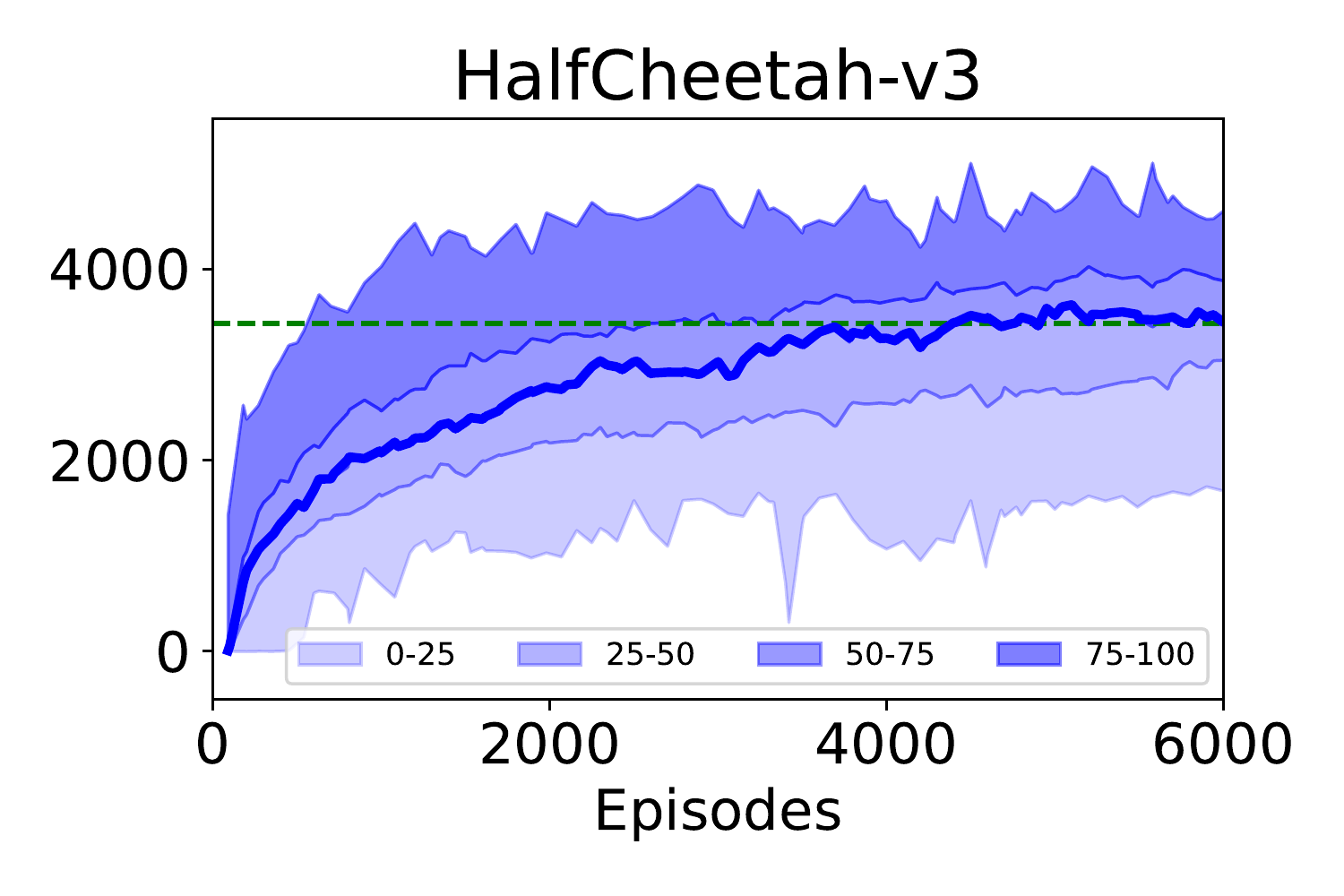}
         \caption*{}
     \end{subfigure}

    \caption{The figures in the top row represent the evaluation of OP-ARS over 100 seeds. The figures in the bottom row represent the evaluation of OP-ARS's sensitivity to hyperparameters. The average reward is plotted against episodes. The thick blue line represents the median curve, and the shaded region corresponds to the percentiles mentioned in the legend.}
        \label{fig:100_seeds_hyper_sens_ars}
\end{figure*}

We compare ARS and TRES with our proposed off-policy variants
on benchmark MuJoCo ~\cite{todorov2012mujoco} tasks available in OpenAI gym ~\cite{gym}.  
%
In this section, we demonstrate that these variants (henceforth, called OP-ARS and OP-TRES) i.) take significantly less number of interactions to reach rewards thresholds, and ii.) are robust to random seeds and hyperparameter choices. Separately, in Appendix \ref{subsec:lqr}, we show that  OP-ARS outperforms ARS even on the Linear Quadratic Regulator (LQR) problem.

\subsection{Sample efficiency}
%
    

    

    



    

Because an ES method is stochastic, the intermediate policies that are learned will be different on each run. Hence, to have a robust comparison, we initiate all random number generators and OpenAI gym environments as a function of a single seed, and then study the performance of our algorithms for eight\footnote{The corresponding numbers in \cite{ARS} and \cite{TRES} are three and six, respectively.} uniformly random seeds. Further, after every ten iterations of our Algorithm, we compute the value function of the current policy (e.g., the policy parameterized by $M_j$ in Algorithm~\ref{alg:offars}) by averaging the total reward obtained over 100 trajectories.
%
%
The sample complexity of our ES methods on each of the eight seeds is the first time the value function of the learned policy is above a certain reward threshold. We use the same thresholds that were used in ~\cite{ARS}. The comparison of sample complexity estimates for ARS, TRES, OP-ARS, and OP-TRES is given in Table~\ref{tab:table_offars_offtres}, Figures~\ref{fig:box_plots}, and \ref{fig:all_seeds}.

In Table~\ref{tab:table_offars_offtres}, the `Env.' and `Th.' columns represent the environment names and its corresponding thresholds. The median of sample complexity estimates is provided under the column titled `Intx.' for different $N$ and $b$ choices. In particular, the values in round brackets are for the case where $b$ random directions are chosen and data from all the $b$ directions is used for updating the policy parameter. The ones outside are for the case where $N$ directions are chosen and data from only the top performing $b$ directions is used for parameter update. Similarly, the numbers under R denote the number of seeds on which the thresholds were reached. Finally, the numbers under \% denote the percentage of data required by OP-ARS (resp. OP-TRES) to reach the threshold compared to ARS (resp. TRES). Clearly, the median estimates (see bold text in Table \ref{tab:table_offars_offtres}) for our variants are significantly lower than those of ARS and TRES.



In Table~\ref{tab:table_offars_offtres}, observe that there are two scenarios: either sampling only $b$ directions and using all of them for parameter update is better (see HalfCheetah and Humanoid in ARS, and HalfCheetah and Ant in TRES) or sampling $N$ directions and then picking the top-performing $b$ directions is better (e.g., Swimmer, Hopper, Walker2d, and Ant in ARS). The former scenario corresponds to the case when the overhead of evaluating the additional $N - b$ directions is costlier than the benefit of exploring more directions and using the best performing ones. In both scenarios, by circumventing this overhead issue via off-policy ranking, OP-ARS and OP-TRES significantly cuts down on sample complexity. Note that Humanoid-v3 is considered the most challenging MuJoCo environment.

In Ant-v3, our experiments confirm that OP-ARS often gets stuck at a local optima. We believe this happens because, near a local maxima, the approximate off-policy ranking forces the updates in different iterations to be in opposing directions. Interleaving with on-policy estimations in such situations appears to help overcome this oscillatory behavior. If our above understanding is correct, then the reason for the stagnation is a bit of both sub-optimal ranking criteria and a poor exploration strategy. We believe additional studies will be needed to develop a robust strategy to deal with this issue.

While Table~\ref{tab:table_offars_offtres} shows that the median of the timesteps needed for the first crossover happens much earlier in our off-policy variants, it doesn't fully capture the variance. We address this in Figures~\ref{fig:box_plots} and \ref{fig:all_seeds}.
%
Figure \ref{fig:box_plots} consists of box plots of the number of interactions required to reach the threshold in ARS, TRES, OP-ARS and OP-TRES. The number of seeds on which each method reached the threshold is mentioned next to the algorithm's name. Clearly, the variance in our approach is either less or comparable to original algorithms in all environments.
The advantage is particularly significant for OP-TRES in Ant-v3 and for OP-ARS in Humanoid-v3.

In Figure~\ref{fig:all_seeds}, the overall progress of various algorithms is shown for different seeds. 
%
In particular, the horizontal green dotted lines represent the specific reward thresholds, while the different curves correspond to different seeds. The blue and red dots on the curves represent the first timestep where the threshold is reached. Finally, the star represents the median of all the dots marked over various runs. The plots clearly show that our methods reach the threshold significantly faster than the original methods in most cases. 

Due to space constraints we have represented the figures only from few environments, the figures of all the remaining environments can be found in Appendix~\ref{appendix:all_plots}. 
The details about the environment is given in Appendix \ref{subsec:env_details} and the implementation details are given in Appendix~\ref{subsec:imp_details}.

\subsection{Random Seeds and Hyperparameter choices}
\label{subsec:random_seeds_and_hyper_choice}
%
    

Previous experiments showed that our methods perform better than ARS and TRES over eight seeds. In this section, we discuss the robustness of OP-ARS to random seeds and hyperparameter choices.

The top row of Figure \ref{fig:100_seeds_hyper_sens_ars} shows the performance of OP-ARS over 100 random seeds sampled from $[0,10000].$ We see that the median (thick blue line) crosses the threshold in most of the environments, demonstrating robustness to seeds.  


Next, we discuss the sensitivity of OP-ARS to the two new hyperparameters we introduce: $h$ (the bandwidth choice in kernel approximation) and $n_b$ (the number of trajectories generated using the behaviour policy). For this, we first identify multiple choices for $h$ and $n_b,$ wherein the the performance of the proposed algorithms is reasonable (see Appendix \ref{appendix:hyperparams}). Next, we run our algorithm under all possible combinations of these hyperparameter choices. The performance plots of this experiment  (see bottom row of Figure~\ref{fig:100_seeds_hyper_sens_ars}) match those obtained in the 100 seed test done above, thereby demonstrating that the algorithm's performance is not sensitive to hyperparameter choices. 
%
We don't look at the sensitivity of OP-ARS to other hyperparameters such as $\nu, \alpha, N, b.$ Such a study in the context of ARS has already been done in ~\cite{ARS} and the role of these parameters in both ARS and our off-policy variant are similar.

\section{Conclusion}

This work proposes an off-policy ranking idea for improving sample efficiency in evolutionary RL methods. While traditional off-policy methods are not directly applicable to deterministic policies, we enable it using kernel approximations.
Our experiments show that our proposed ARS and TRES variants have roughly the same run time as the original, but 
reach reward thresholds with only 50-80\% as much interactions. We believe our approach is easily extendable to other ES or Random Search methods as well. A promising direction of future work would be to investigate the same theoretically.
Separately, recent hybrid algorithms that mix ES and Deep RL methods have shown to be more sample efficient than ES methods,
for example, CEM-RL~\cite{CEM-RL}.
We strongly believe that our off-policy ranking idea can help in these hybrid algorithms as well. 

\appendix

\onecolumn

\setcounter{table}{0}
\renewcommand{\thetable}{A\arabic{table}}
\setcounter{figure}{0}
\renewcommand{\thefigure}{A\arabic{figure}}
\setcounter{equation}{0}
\renewcommand{\theequation}{A\arabic{equation}}
\setcounter{algorithm}{0}
\renewcommand{\thealgorithm}{A\arabic{algorithm}}

\section{Differences between ARS and OP-ARS}
\label{appendix:ars_vs_ours}

The Section~\ref{s:Method} of the main paper provides a detailed explanation of our proposed idea. In particular, it describes how we apply our technique on ARS to get OP-ARS. In this section, we present the exact differences between the proposed OP-ARS (given in Algorithm\ref{alg:offars}) and the original ARS algorithm. Table~\ref{tab:ars_vs_ours} indicates the exact steps that differ between the algorithms and its implications.

\begin{table}[h!] 
\centering
\begin{tabular*}{\linewidth}{p{0.04\linewidth}p{0.25\linewidth}p{0.25\linewidth}p{0.36\linewidth}} 
 \toprule
 Step & OP-ARS & ARS &  Implications\\
 \midrule
 6 & run $n_b$ trajectories & there is no corresponding step in ARS &  the trajectories are run using policy parameterized by $M_j$ (used as behavior data for off-policy evaluation)\\
 8 & collects only $\mathbf{2b}$ rollouts (note that $b<N$) & step 5 in ARS; collects $\mathbf{2N}$ rollouts &  we collect data required only for update step\\
 7 & we sort $\delta_k$ based on the fitness function \eqref{eq:fpifromdeltak} derived using off-policy technique & step 6 in ARS; sorts $\delta_k$ based on returns from the $2N$ trajectories &  we generate less data as we find the approximate ranking of $\delta$s using off-policy technique even before generating trajectories, unlike in ARS where trajectories are run for all $2N$ policies to find the rankings\\
 9 & we use $\hat{\eta}(\pi)$ to denote the total reward received from policy $\pi$ & step 7 in ARS; uses $r(\pi)$ for the total reward received from policy $\pi$ &  we use $\hat{\eta}(\cdot)$ to denote the total reward because we use $r(\cdot)$ for one-step reward\\
 7, 8 & we first rank; then collect only data from $2b$ policies & step 5, 6 in ARS; first collect data from $2N$ policies, then rank &  we generate less samples as mentioned earlier\\
 \bottomrule
\end{tabular*}
\caption{\label{tab:ars_vs_ours}Key differences between ARS and OP-ARS. The step column refers to the step number in Algorithm \ref{alg:offars}.}
\end{table}

\section{Review of TRES}
\label{tres_review}
The Section\ref{sebsec:tres_overview} of the main paper provides a quick overview of Trust Region Evolutionary Strategies(TRES)\cite{TRES}. In this section, we provide a detailed review of it.

Similar to ARS, TRES also solves the optimization problem mentioned in \eqref{e:Opt.Problem}. As discussed in Section~\ref{sebsec:ars_review}, in general this objective function need not be smooth. In order to tackle this issue, the authors of TRES optimize the guassian smoothed version of the above objective function. \cite{TRES} use antithetic ES gradient estimator given in \cite{ES} to perform stochastic gradient ascent on the smoothed objective function. The antithetic ES gradient estimator is given by:
\begin{equation}
\label{app:eq.grad}
    \hat{\nabla}_N \eta(\pi_\theta) = \frac{1}{2N\sigma^2}\sum_{i=1}^N((\eta(\pi_{x_i}) - \eta(\pi_{2\cdot\theta - x_i}))(x_i - \theta))
\end{equation}
where $x_i \sim \mathcal{N}(\theta, \sigma^2I)$.

In every iteration of ES \cite{ES} algorithm, search parameters $\{x_1, x_2, ... , x_N\}$ are sampled from $\mathcal{N}(\theta,\sigma^2I)$ and evaluated by performing rollouts. The data generated from these rollouts is used to update the current parameters using the gradient computed using \eqref{app:eq.grad}. 
However, for the next iteration, the data is again generated to compute the gradient. The authors of TRES improve upon this drawback by proposing a surrogate objective function which can reuse the data from these rollouts to update the parameters multiple times. 
We next show how the surrogate objective function is developed by \cite{TRES}. Going forward, we slightly abuse the notation for $\eta(\pi_\theta)$ as $\eta(\theta)$.

\cite{TRES} propose a local approximation $L_\theta(\tilde{\theta})$ to $\eta(\pi_{\tilde{\theta}})$. Such an approximation will help optimize for $\eta(\pi_{\tilde{\theta}})$ using off-policy data generated from $\pi_\theta$ instead of $\pi_{\tilde{\theta}}$.
The proposed local approximation is given by:
\begin{equation}
\label{app:eq.TRES.L}
   L_\theta(\tilde{\theta}) = \eta(\theta) + \sum_{i=0}^{d-1}\sum_{x^{1...i}}p_\theta(x^{1...i})\sum_{x^{i+1}}p_{\tilde{\theta}}(x^{i+1})A_{\theta}(x^{1...i+1})
\end{equation}
where, $X^{1...i}$ represents the first $i$ dimensions of $X$ (the $d$-dimensional search parameter), $p_\theta(x^{1...i})$ represents the marginal distribution of $X^{1...i}$ for $i \in \{1,...,d\}$. \cite{TRES} show that, any $\theta$ that improves $L_\theta(\tilde{\theta})$ improves $\eta(\pi_{\tilde{\theta}})$. They also derive the bound between the proposed $L_\theta(\tilde{\theta})$ and $\eta(\pi_{\tilde{\theta}})$ (see \cite[Theorem 1]{TRES}) as:

\begin{equation}
\label{app:eq.TRES.L.bound}
   | \eta(\tilde{\theta}) - L_\theta(\tilde{\theta}) | \leq 2\epsilon d (d+1) \alpha^2
\end{equation}
where $\epsilon = \max_{x^{1...i}} |A_{\theta}(x^{1...i}|$ and $\alpha = D_{TV}^{max}(\theta,\tilde{\theta})$.
Next, they show that by maximizing $L_\theta(\tilde{\theta}) - 2\epsilon d (d+1)  D_{KL}^{max}(\theta,\tilde{\theta})$, the actual objective $\eta(\tilde{\pi})$ is non-decreasing. The above problem can now be rephrased as a constrained optimization problem in each iteration as: $\max_{\tilde{\theta}} L_\theta(\tilde{\theta})$ subjected to $D_{KL}^{max}(\theta,\tilde{\theta}) \le \delta$. By replacing the summations to expectations and by using importance sampling ratio, the final objective function is written as(refer \cite[24]{TRES}):
\begin{equation*}
    \max_{\tilde{\theta}} \bE_{X\sim\theta}\left[\sum_{i=0}^{d-1}\frac{p_{\tilde{\theta}}(X^{i+1})}{p_{{\theta}}(X^{i+1})}V_\theta(X^{1...i+1}) \right]
    subject~to~ D_{KL}^{max}(\theta,\tilde{\theta}) \le \delta
\end{equation*}
where, $V_\theta(x^{1...i}) = \bE_{X^{i+1...d}\sim \theta}\left[ \eta(X) | X^{1...i} = x^{1...i} \right]$.
The above constrained optimization problem is hard to solve, hence, the authors take inspiration from PPO \cite{PPO} to propose a surrogate objective function with clipped probability ratio given by:

\begin{equation}
\label{app:eq.TRES.clip_obj}
    L_{\theta}^{CLIP}(\tilde{\theta}) = \bE_{X\sim\theta}\left[ \sum_{i=1}^d min(l_i(\tilde{\theta}), l_i^{CLIP}(\tilde{\theta}))\right]
\end{equation}
where, $r_i(\tilde{\theta}) = \frac{p_{\tilde{\theta}}(X^i)}{p_\theta(X^i)}$ ,$l_i(\tilde{\theta}) = r_i(\tilde{\theta})V_\theta(X^{1...i})$ and $l_i^{CLIP}(\tilde{\theta}) = clip(r_i(\tilde{\theta}), 1-\lambda, 1+\lambda)V_\theta(X^{1...i})$

The above surrogate objective is the final practical objective function optimized in every iteration of TRES. In each iteration of TRES, the gradient of above is computed and used to update the parameters multiple times. The distributed practical implementable TRES algorithm is given in \cite[Algorithm 1]{TRES}.

\section{OP-TRES}
\label{op_tres}

The Section~\ref{s:Method} of the main paper discusses the derivation of our fitness function $f_{\pi_j}(\delta_k,h)$ (see \eqref{eq:fpifromdeltak}). Since, TRES also uses deterministic linear policies, the same fitness function derived in Section~\ref{s:Method} can be used to rank the directions in OP-TRES. Hence, we first rank the directions using \eqref{eq:fpifromdeltak}), and then continue with the generation of rollouts for the $b$ top-performing directions. The proposed OP-TRES method is given in Algorithm~\ref{alg:offtres}.

\begin{algorithm}[h]
  \caption{Off-policy TRES}
  \label{alg:offtres}
\begin{algorithmic}[1]
  \STATE {\bfseries Require:} noise standard deviation $\sigma$, initial policy parameter $\theta_0$, epoch number K, learning rate $\alpha$, clip factor $\lambda$, bandwidth to use for kernel approximation $h$, number of behaviour policy trajectories to run $n_b$, $N$ number search directions to sample in each iteration, $b$ number of top-performing directions to consider while updating the parameters
  \STATE {\bfseries Ensure:} $N$ workers with known random seeds, initial parameters $\theta_0$
  \FOR{each iteration t = 0,1,2,...}
  \STATE Sample $x_i \sim \mathcal{N}(\theta_t,\sigma^2I)$ for $i=1,...,N$
  \STATE  Run $n_b$ number of trajectories using policy parameterized by $\theta_t$, resulting in $N_d$ number of interactions
  \STATE  Sort the directions $\delta_i$ based on $f_{\pi_{t}}(\delta_i, h)$ scores (using \eqref{eq:fpifromdeltak}), where $\delta_i = x_i - \theta_t$
  \STATE Compute rollout returns of the $b$ top-performing directions
  \STATE Send all rollout returns to every worker
  \FOR{each worker $i$= 1,...,N}
    \STATE Reconstruct all search parameters
    \STATE Compute value functions with rollout returns
    \STATE Let $\theta_{t,1} = \theta_t$
    \FOR{each epoch $j$=1,...,K}
        \STATE Reusing the sampled data from $\theta_t$
        \STATE $\theta_{t,j+1} = \theta_{t,j} + \alpha \nabla L_{\theta_t}^{CLIP}(\theta_{t,j})$
    \ENDFOR
    \STATE Update policy parameters via $\theta_{t+1} = \theta_{t,K+1}$
  \ENDFOR
 \ENDFOR
   
\end{algorithmic}
\end{algorithm}

\section{Environment details}

In this section we briefly discuss the environment details. We first start off with OpenAI gym \cite{gym} and MuJoCo \cite{todorov2012mujoco}. Later, we move on to Linear Quadratic Regulator (LQR).

\label{subsec:env_details}

\subsection{MuJoCo and OpenAI Gym}

Open AI gym is an open-source python library used in benchmarking reinforcement learning algorithms. 
It contains environments ranging from simple(like pendulum) to very complex setups(like robot locomotion and atari games).
OpenAI gym abstracts these environments and provide an easy API to train RL agents. The agents observe the current state of the environment and apply an action accordingly to get the instantaneous reward. 
The robotic locomotion tasks like (HalfCheetah-v3, Hopper-v3, Humanoid-v3, etc.) use the MuJoCo simulator in the backend. Notably, the MuJoCo robotic locomotion tasks consist of continuous state-space and action-space. The state representation in each of these environments is the positions and velocities of the center of mass and various robot joints. The action space is the torque to be applied at each of the robot's joints. The instantaneous reward is based on the distance travelled by the robot and the control cost due to the action taken.

\subsection{Linear Quadratic Regulator(LQR)}
\label{appendix:lqr_env}

Linear Quadratic Regulator is a linear dynamical system with quadratic cost. The goal is to devise a control strategy that minimizes the cost. Mathematically, it can be written as the following:
$$\min_{u_0, u_1,...} \lim_{T \rightarrow \infty} \frac{1}{T} \mathbb{E}[\Sigma_{t=0}^{T-1} x_t^T Q x_t + u_t^T R u_t]$$
$$s.t. x_{t+1} = Ax_t + Bu_t + w_t$$

~\cite{ARS} use the LQR instance defined by ~\cite{lqr_def}. Here,
$$A = \begin{bmatrix}
1.01 & 0.01 & 0\\
0.01 & 1.01 & 0.01\\
0 & 0.01 & 1.01
\end{bmatrix}
B = I, Q= 10^{-3}I, R = I$$
We use the same system mentioned above for our experiments in this paper.

\section{Implementation details}
\label{subsec:imp_details}

We adopt the code from \cite{ARS} and modify it accordingly to implement our off-policy ranking. ~\cite{ARS} implemented a parallel version of the ARS algorithm using the python package Ray ~\cite{ray}. They create a centralized noise table with standard normal entries and pass on the starting indices between workers instead of sending the entire perturbation $\delta$. This centralized table avoids the bottleneck of communication. 
Another key point is that, at every time step, MuJoCo locomotion tasks have a survival reward if the robot does not fall over. This survival rewards cause the ARS and TRES algorithms to learn policies that stand still for long. Hence, ~\cite{ARS} suggest subtracting the survival reward during the training phase. We also adopt the same. The code for ARS is borrowed from \href{https://github.com/modestyachts/ARS}{https://github.com/modestyachts/ARS} and the code for LQR experiments is borrowed from \href{https://github.com/benjamin-recht/benjamin-recht.github.io/tree/master/secret\_playground}{https://github.com/benjamin-recht/benjamin-recht.github.io/tree/master/secret\_playground}.

\section{Hyperparameters}
\label{appendix:hyperparams}

In this section, we would like to describe the set of hyperparameters used in our experiments. The ARS and TRES algorithms have a predefined set of hyperparameters, which have been fine-tuned in the corresponding papers. Hence, we use the same hyperparameters in our algorithms for most of the environments. Our off-policy variants OP-ARS and OP-TRES have two new hyperparameters: the number of trajectories to run using behavior policy $n_b$ and the bandwidth in kernel function $h$. We try different values for $n_b$ and $h$ as mentioned in Table~\ref{tab:all_hypers} to do hyperparameter search. The final set of hyperparameters that gave us the best results are mentioned in Table~\ref{tab:final_hypers}.


\begin{table*}[h]
\centering
\begin{tabular}{|c|cccc|cccc|}
\hline
\multirow{2}{*}{Env.(-v3)}& \multicolumn{4}{c|}{OP-ARS} & \multicolumn{4}{c|}{OP-TRES}\\
\cline{2-9}
& N & b & $n_b$ & h  & N & b & $n_b$ & h \\
\hline
 Swimmer & 2 & 1 & 1,2 & 1.0,0.5,0.25,0.1  & 4 & 2 & 1,2 &1.0,0.5,0.25,0.1  \\
 Hopper & 8 & 4 & 1,2 & 1.0,0.5,0.25,0.1  & - & - & - & - \\
 HalfCheetah & 32 & 4 & 1,2 & 1.0,0.5,0.25,0.1  & 16 & 8 & 1,2 &1.0,0.5,0.25,0.1 \\
 Walker2d & 40 & 30 & 1,2 & 1.0,0.5,0.25,0.1  & - & - & - &-\\
 Ant & 60 & 20 & 1,2 & 1.0,0.5,0.25,0.1  & 40 & 20 & 1,2 &1.0,0.5,0.25,0.1 \\
 Humanoid & 350 & 230 & 1,2 & 1.0,0.5,0.25,0.1  & - & - & - &-\\
\hline\end{tabular}
\caption{\label{tab:all_hypers}Hyperparameter grid used in each environment}
\end{table*}

\begin{table*}[h]
\centering
\begin{tabular}{|c|cccc|cccc|}
\hline
\multirow{2}{*}{Env.(-v3)}& \multicolumn{4}{c|}{OP-ARS} & \multicolumn{4}{c|}{OP-TRES}\\
\cline{2-9}
& N & b & $n_b$ & h  & N & b & $n_b$ & h \\
\hline
 Swimmer & 2 & 1 & 2 & 0.1  & 4 & 2 & 1 & 0.1  \\
 Hopper & 8 & 4 & 2 & 0.25  & - & - & - & - \\
 HalfCheetah & 32 & 4 & 2 & 1.0 & 16 & 8 & 1 & 0.5 \\
 Walker2d & 40 & 30 & 2 & 0.5  & - & - & - &-\\
 Ant & 60 & 20 & 1 & 0.25 & 40 & 20 & 1 & 0.5 \\
 Humanoid & 350 & 230 & 2 & 0.25 & - & - & - &-\\
\hline\end{tabular}
\caption{\label{tab:final_hypers}Best performing hyperparameters used to generate results in Table \ref{tab:table_offars_offtres} and Figures  \ref{fig:box_plots} and \ref{fig:all_seeds}}
\end{table*}

\section{LQR experiments}
\label{subsec:lqr}
As mentioned in Section ~4.3 of ~\cite{ARS}, MuJoCo robotic tasks have some drawbacks. Most importantly, the optimal policies of these environments are unknown. Therefore, one is unsure how their algorithm's learned policy compares to the optimal policy. One good idea is to apply the algorithms to simple well known, and well-studied environments whose optimal policy is known. ~\cite{ARS} chose Linear Quadratic Regulator (LQR) with unknown dynamics for this benchmarking. More details about the environment are given in Appendix \ref{appendix:lqr_env}. 

\begin{figure*}[h]
     \centering
     \begin{subfigure}[c]{0.3\textwidth}
         \centering
         \includegraphics[width=\textwidth]{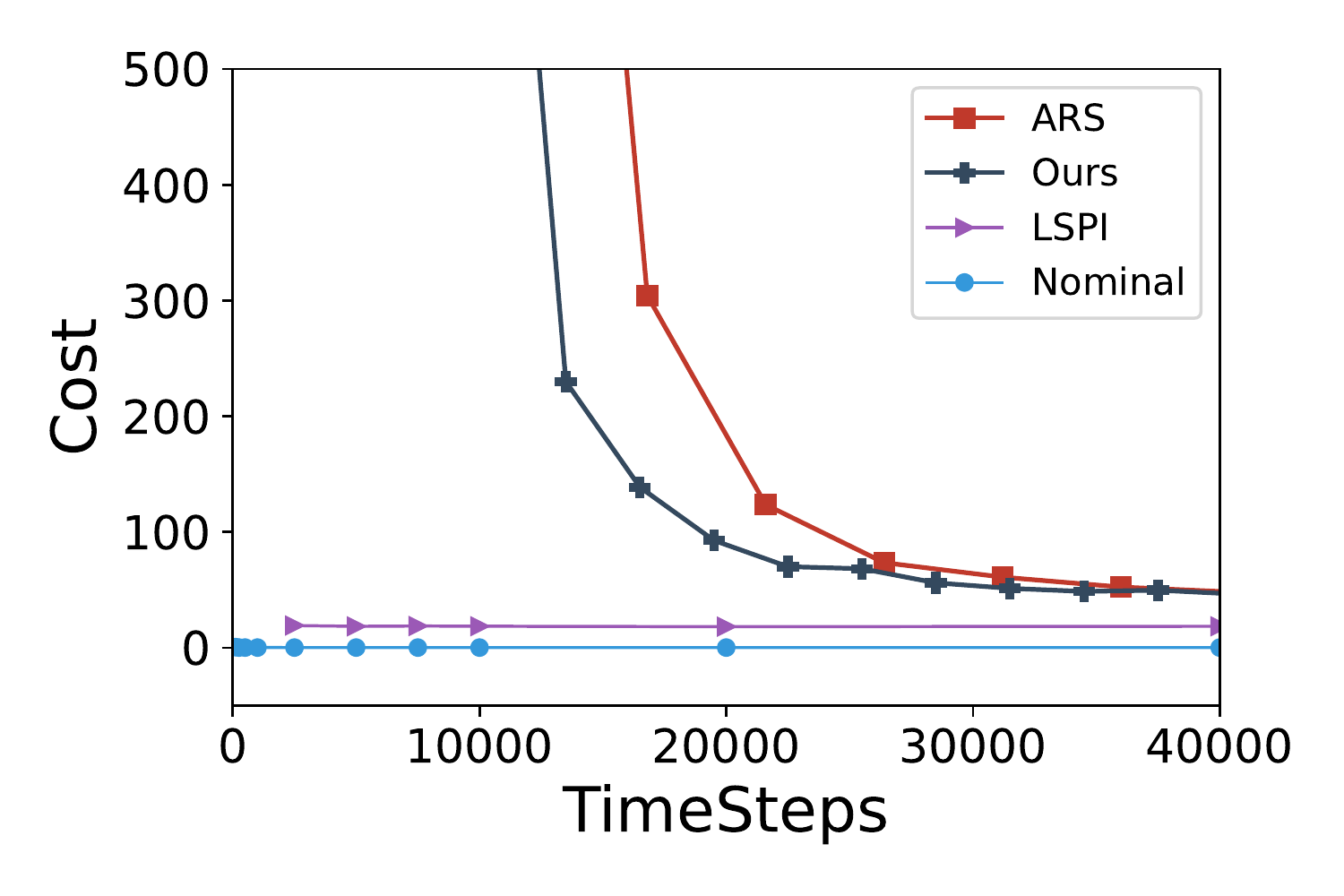}
         \caption{Cost vs TimeSteps}
         \label{fig:lqr_cst_vs_ts}
     \end{subfigure}
     \begin{subfigure}[c]{0.3\textwidth}
         \centering
         \includegraphics[width=\textwidth]{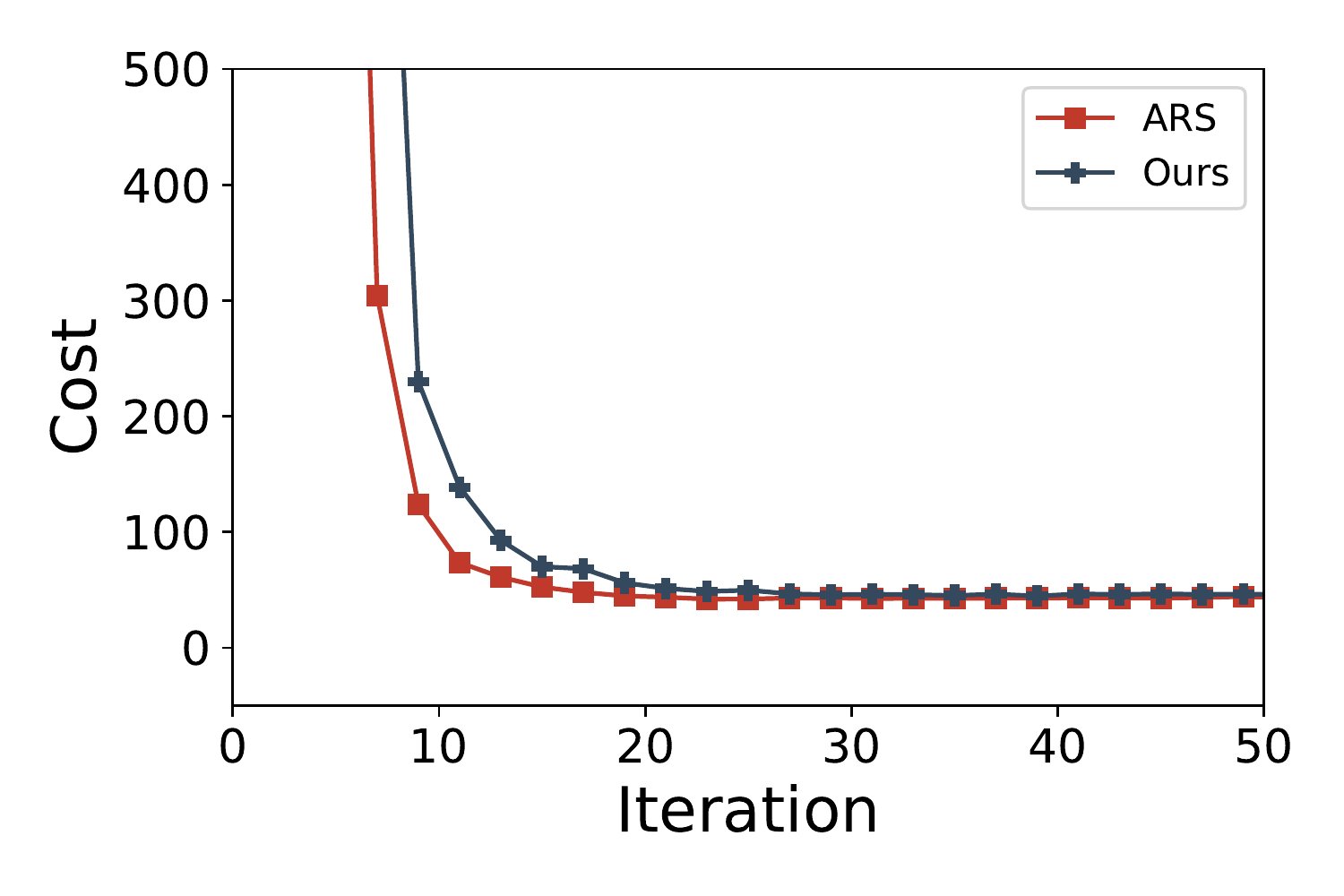}
         \caption{Cost vs Iterations}
         \label{fig:lqr_cst_vs_itr}
     \end{subfigure}
     \begin{subfigure}[c]{0.3\textwidth}
         \centering
         \includegraphics[width=\textwidth]{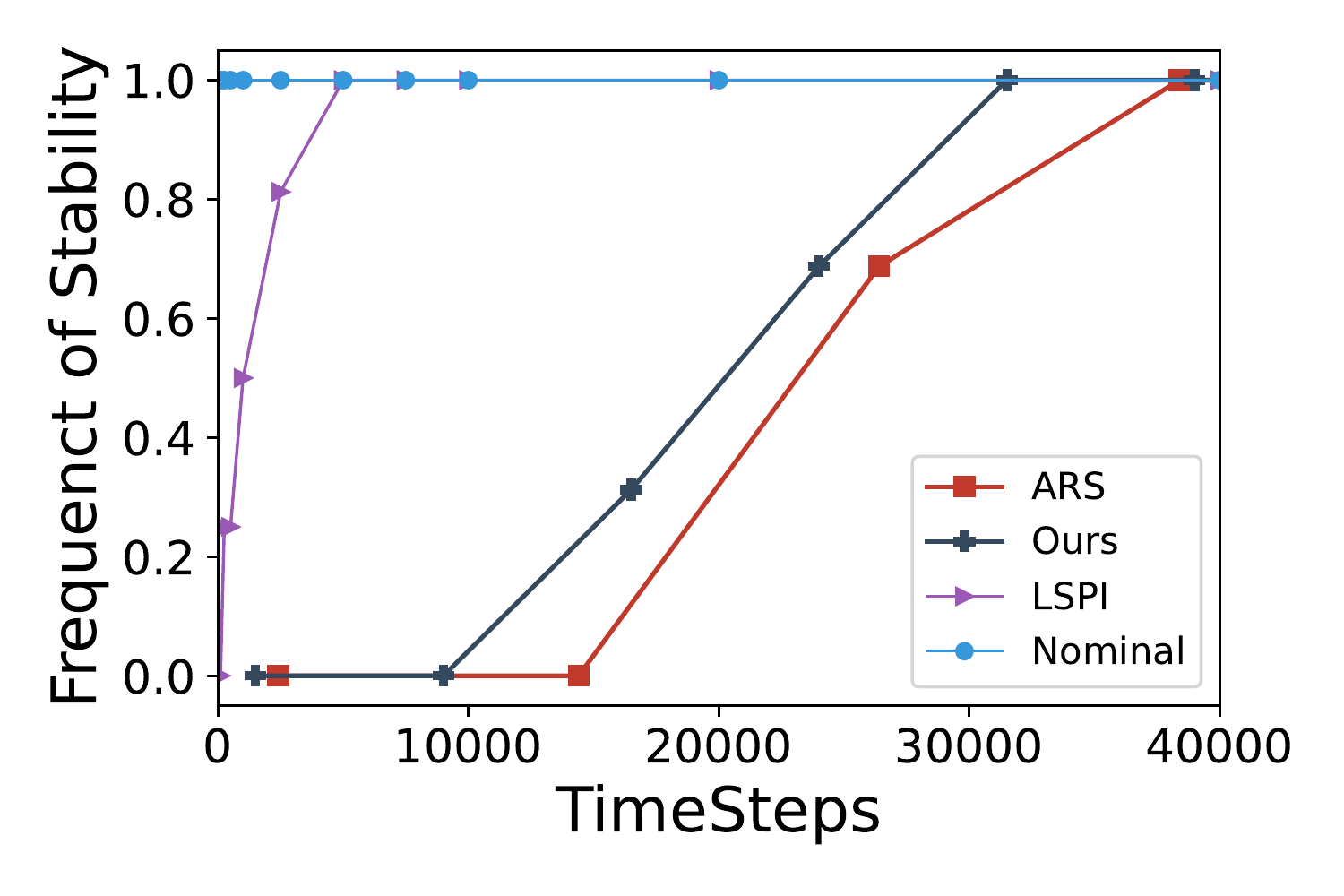}
         \caption{Frequency of Stability}
         \label{fig:lqr_freq_stab}
     \end{subfigure}
        \caption{Comparison of sample efficiency and stability of various algorithms on LQR.}
        \label{fig:lqr}
\end{figure*}

We use the same system used by ~\cite{ARS} to compare our method with model-based Nominal Control, LSPI~\cite{lspi} and ARS~\cite{ARS}. 
As shown by ~\cite{ARS}, the Nominal method is more sample efficient than LSPI and ARS by several orders of magnitude, showing that there is a scope for improvement.
Our experiments show that our method is more sample efficient than ARS as shown in Figure \ref{fig:lqr_cst_vs_ts}
. Figure \ref{fig:lqr_freq_stab} shows that our algorithm is better than ARS in terms of frequency of stability.

\section{Time Comparison}
\label{appendix:time}

In this subsection, we compare the median wall clock time required to reach the specified threshold by each method. Along with ARS and TRES, we also compare our methods with CEM-RL \cite{CEM-RL}. CEM-RL is an hybrid algorithm that mix ES and Deep RL methods. It is more sample efficient than vanilla ES methods. However, their run time is too high. In each iteration of CEM-RL, half of the policies are trained using off-policy learning techniques like DDPG~\cite{DDPG} and TD3~\cite{TD3}, leaving the other half policies untouched; however, in the end, all the policies are evaluated in an on-policy fashion.

Table \ref{tab:app5} provides a quick summary of the median wall-clock time (in seconds) taken by each algorithm to reach the threshold. ARS, TRES, and our method are run on CPU (Threadripper 3990X), while CEM-RL is run on GPU (Nvidia RTX 3080-TI). A critical point to note here is that CEM-RL could not solve the Humanoid-v3 environment, which is considered the most challenging environment. 

While it is a common practice to report the wall-clock time of the algorithm, we believe it might not be a good metric in real-world. The algorithms used in our benchmarking namely, ARS, TRES, OP-ARS, OP-TRES and CEM-RL use multiple copies of same environment in parallel to achieve the mentioned speed. However, in real world, it is usually the case that we work with one robot. Hence, in these cases, measuring the algorithms efficiency based on time required for interactions and time to run the implementation of the algorithms which use a single environment, may be useful.

\begin{table*}[h!] 
\centering

\begin{tabular}{ccccccc} 
 \toprule
 Environment & Threshold & ARS & OP-ARS & TRES  & OP-TRES & CEM-RL \\
 \midrule
 Swimmer-v3 & 325 & 40 & 52 & 32 & 63 & -\\
 Hopper-v3 & 3120 & 48 & 80 & - & - & 5703\\
 HalfCheetah-v3 & 3430 & 42 & 110 & 37 & 37 &  2268\\
 Walker2d-v3 & 4390 & 364 & 366 & - & - & 15363 \\
 Ant-v3 & 3580 & 322 & 665 & 581 & 533 & 16989\\
 Humanoid-v3 & 6000 & 1050 & 625 & - & - & - \\
 \bottomrule
\end{tabular}
\caption{\label{tab:app5}Comparing median wall-clock time (in seconds) of ARS, TRES, OP-ARS, OP-TRES and CEM-RL.}
\end{table*}

\newpage

\section{All Plots}
\label{appendix:all_plots}

Due to the lack of space we presented only few plots in the main paper. In this subsection we add all the remaining plots. The plots are in similar lines to the ones presented in the main paper.

\begin{figure*}[h!]
     \centering
     \begin{subfigure}[c]{0.3\textwidth}
         \centering
         \includegraphics[width=\textwidth]{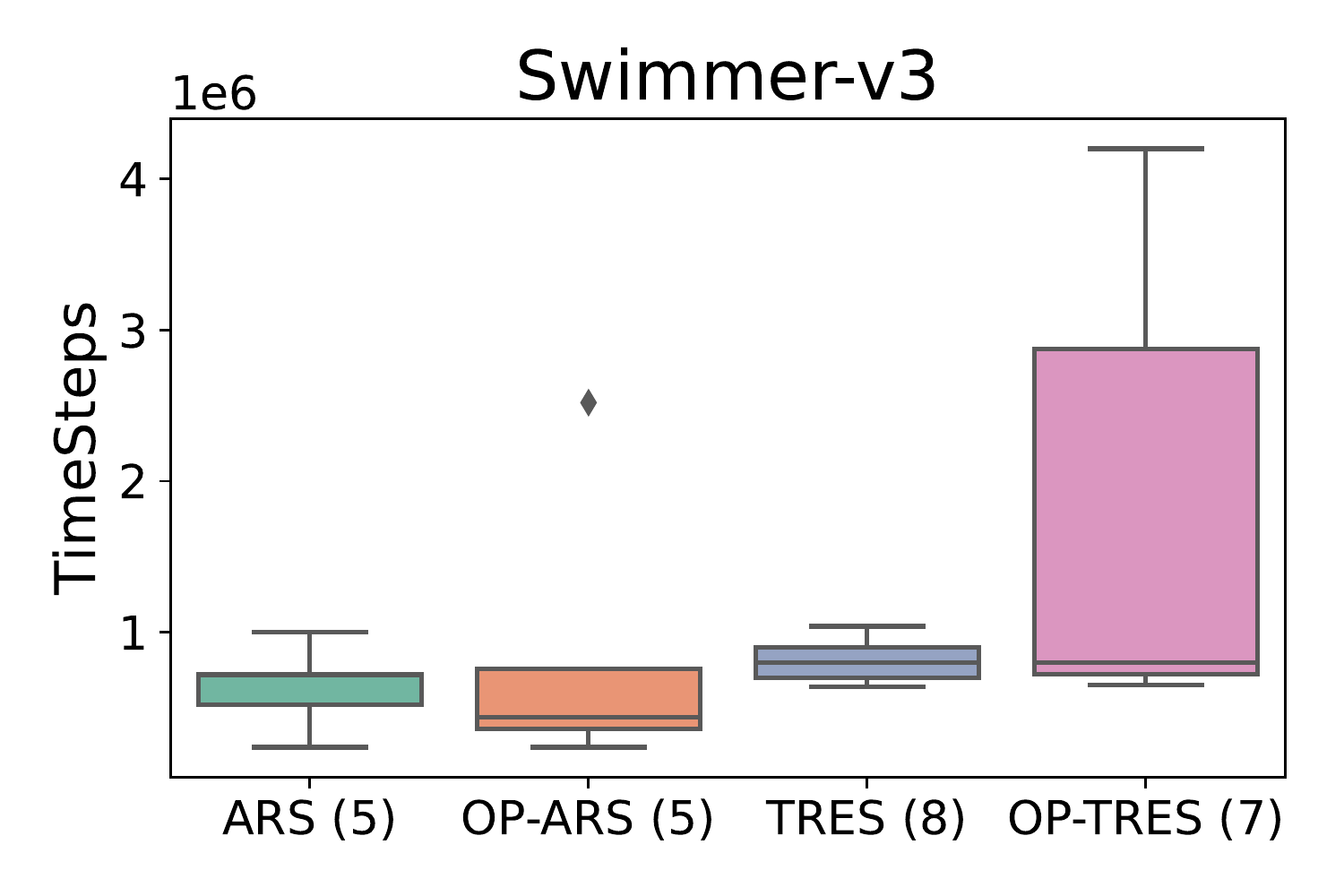}
     \end{subfigure}
     \begin{subfigure}[c]{0.3\textwidth}
         \centering
         \includegraphics[width=\textwidth]{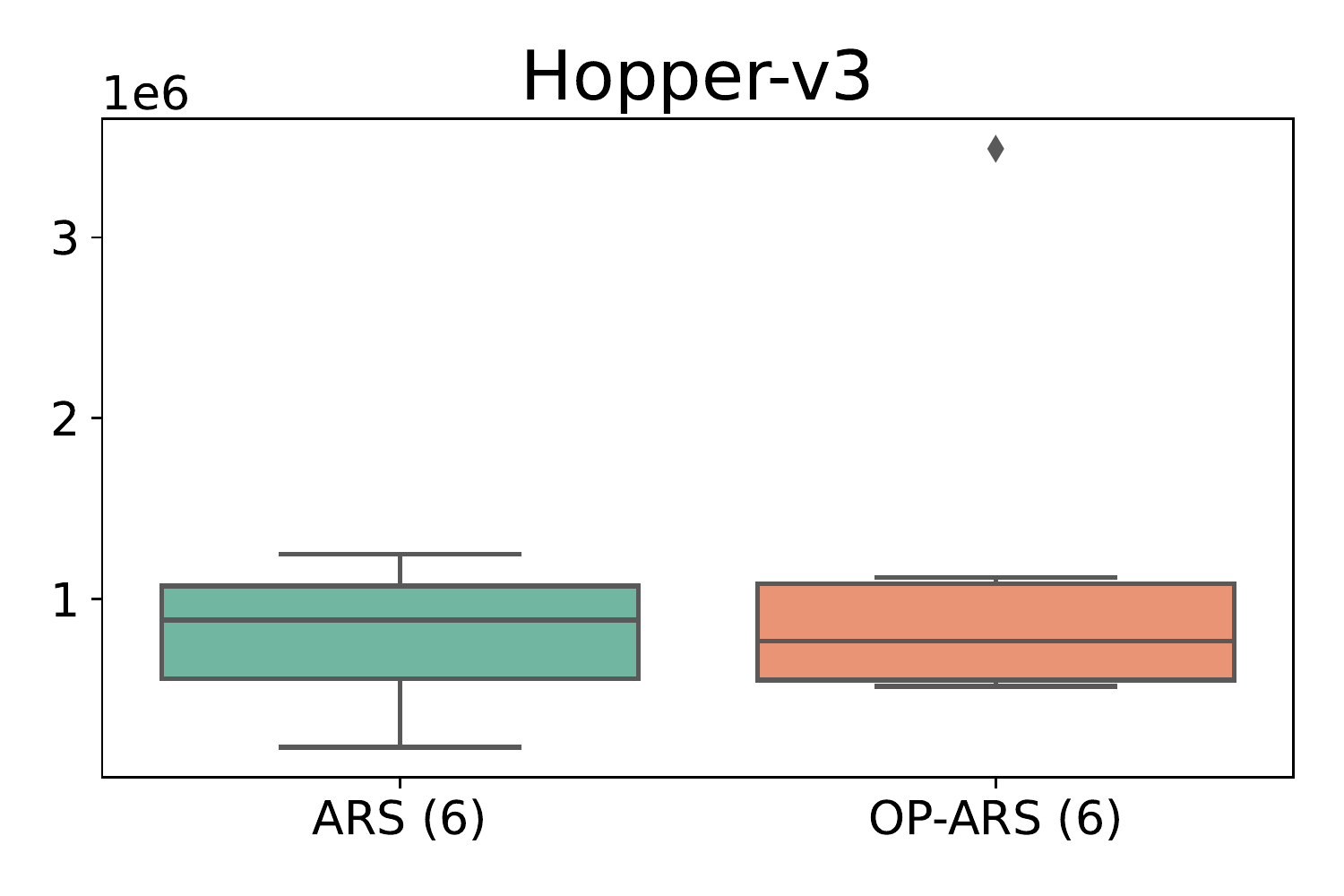}
     \end{subfigure}
     \begin{subfigure}[c]{0.3\textwidth}
         \centering
         \includegraphics[width=\textwidth]{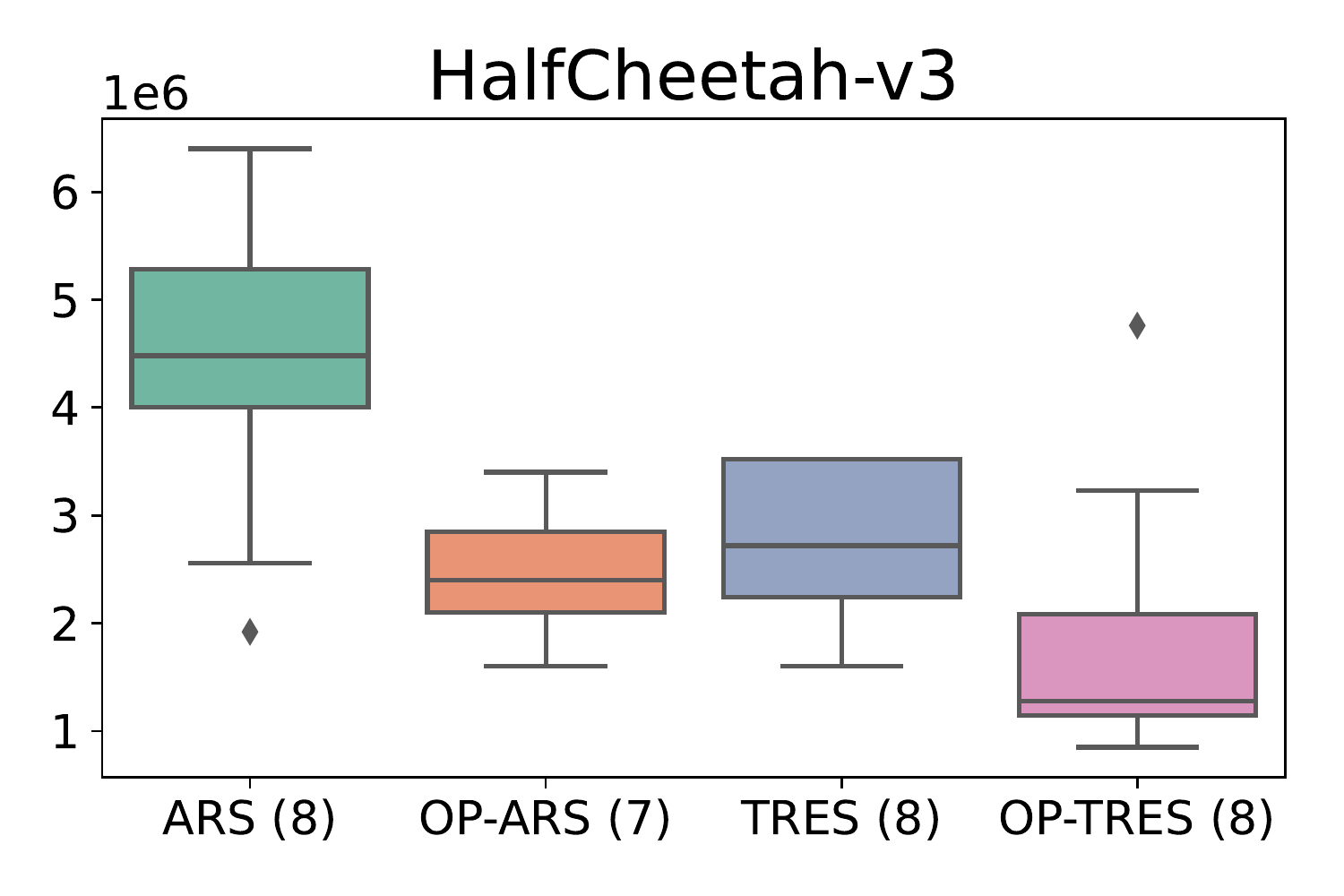}
     \end{subfigure}
     \begin{subfigure}[c]{0.3\textwidth}
         \centering
         \includegraphics[width=\textwidth]{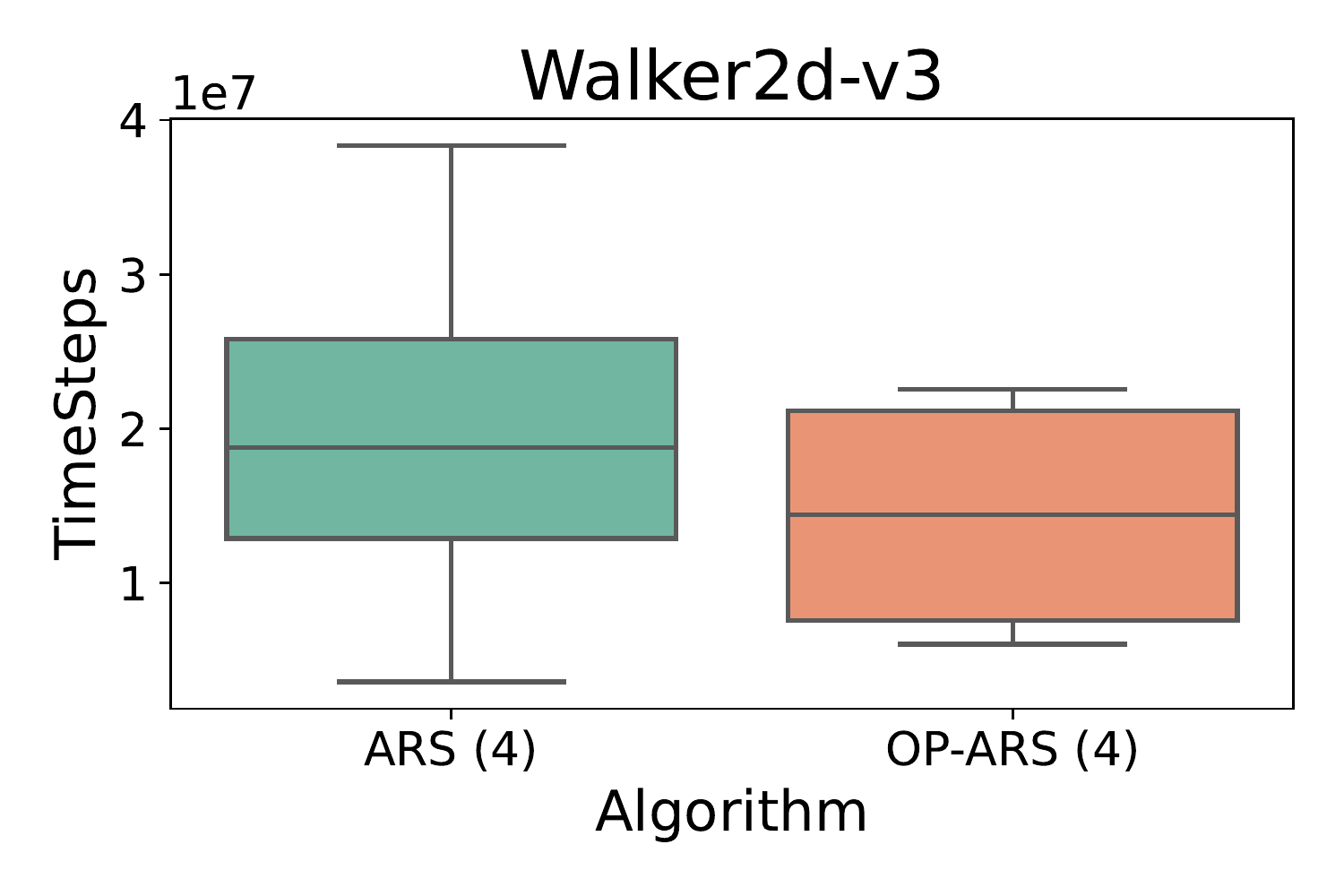}
     \end{subfigure}
     \begin{subfigure}[c]{0.3\textwidth}
         \centering
         \includegraphics[width=\textwidth]{box_plots/Ant-v3.pdf}
     \end{subfigure}
     \begin{subfigure}[c]{0.3\textwidth}
         \centering
         \includegraphics[width=\textwidth]{box_plots/Humanoid-v3.pdf}
     \end{subfigure}
        \caption{Box plots of number of interactions required to reach the reward threshold in ARS, TRES, OP-ARS and OP-TRES. The number next to the algorithm's name represents the number of seeds in which the threshold was reached.}
        \label{fig:appendix.box_plots}
\end{figure*}

Figure \ref{fig:appendix.box_plots} contains box-plots of timesteps required to reach the reward threshold in all the environments. It can clearly be seen that, our proposed method perform the best in this aspect.

\begin{figure*}[h!]
     \centering
     \begin{subfigure}[c]{0.26\textwidth}
         \centering
         \includegraphics[width=\textwidth]{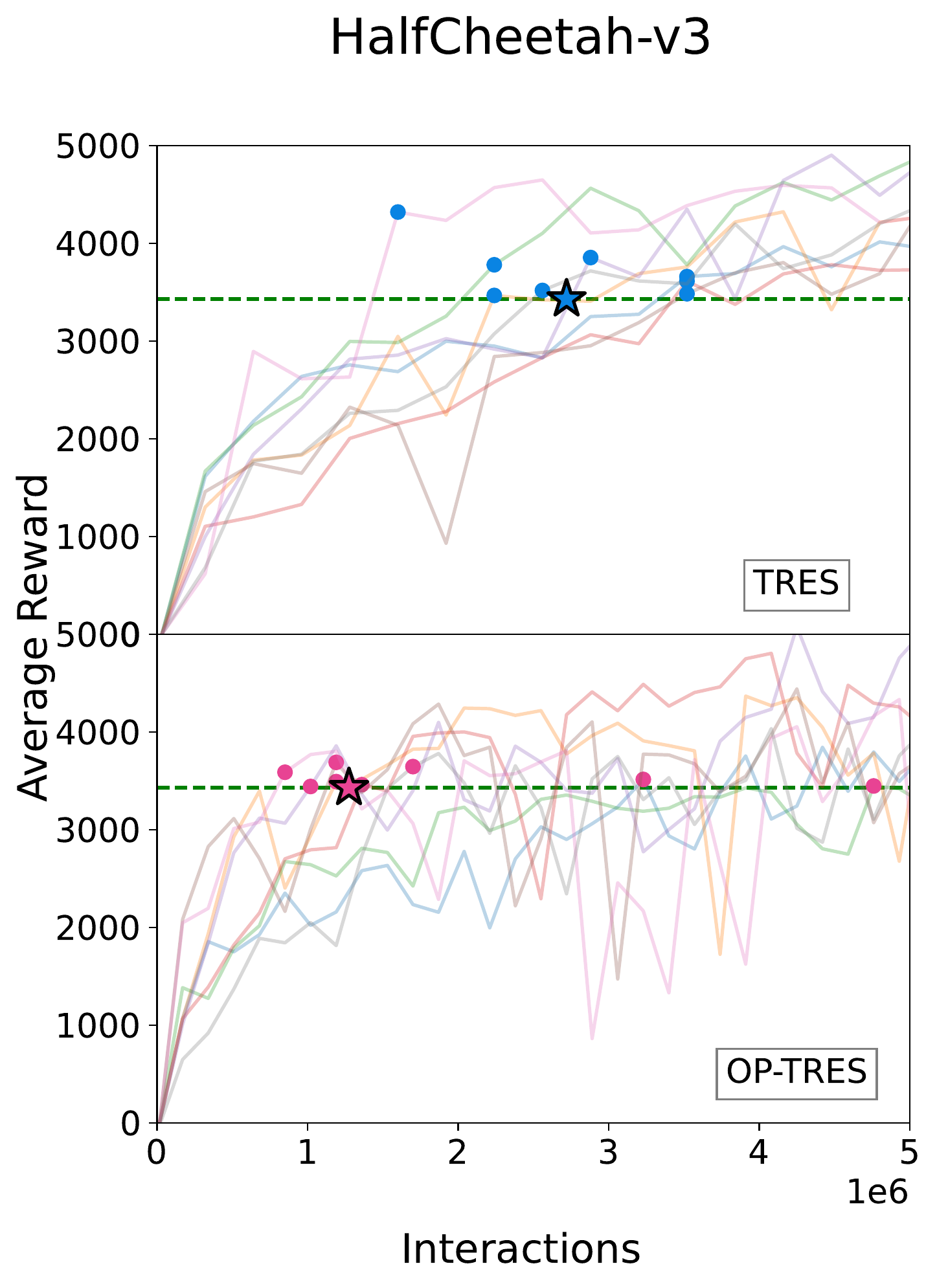}
     \end{subfigure}
     \begin{subfigure}[c]{0.245\textwidth}
         \centering
         \includegraphics[width=\textwidth]{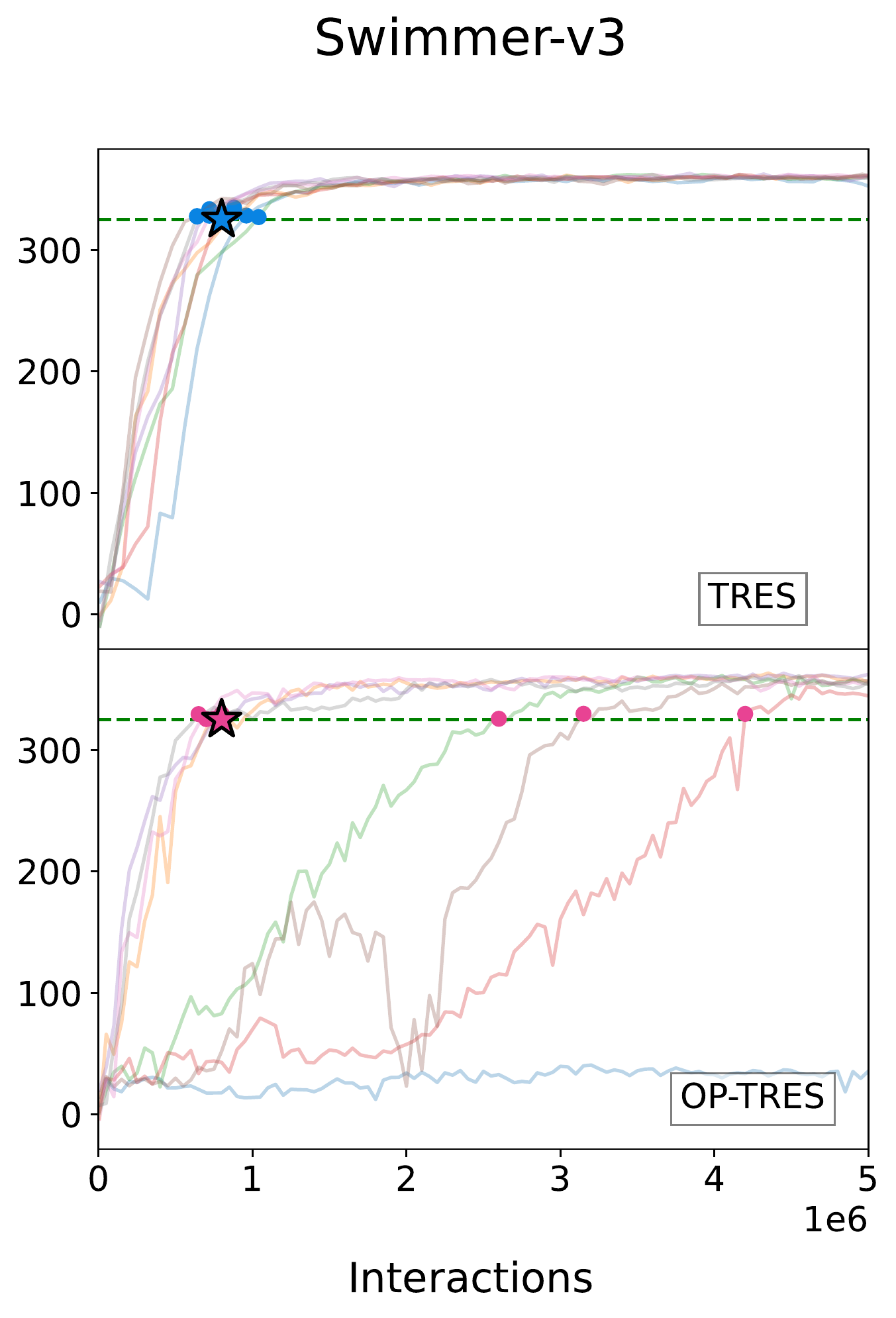}
     \end{subfigure}
     \begin{subfigure}[c]{0.26\textwidth}
         \centering
         \includegraphics[width=\textwidth]{all_seeds_tres/Ant-v3_split.pdf}
     \end{subfigure}
        \caption{Figures representing the trajectories of various runs of TRES and OP-TRES algorithms where the number of interactions with the environment is plotted against the average reward.}
        \label{fig:appendix.all_seeds_tres}
\end{figure*}

\begin{figure*}[h!]
     \centering
     \begin{subfigure}[c]{0.26\textwidth}
         \centering
         \includegraphics[width=\textwidth]{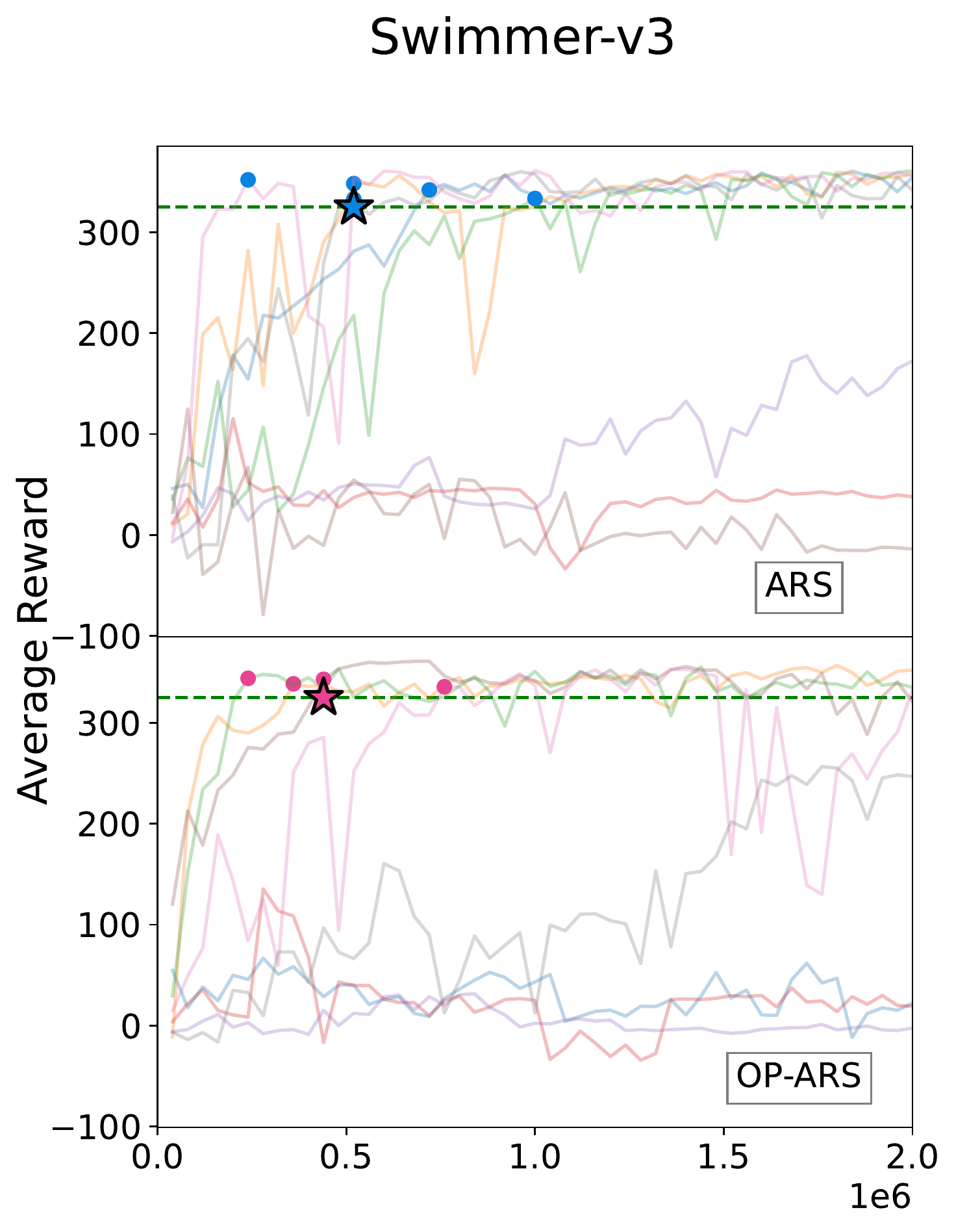}
     \end{subfigure}
     \begin{subfigure}[c]{0.25\textwidth}
         \centering
         \includegraphics[width=\textwidth]{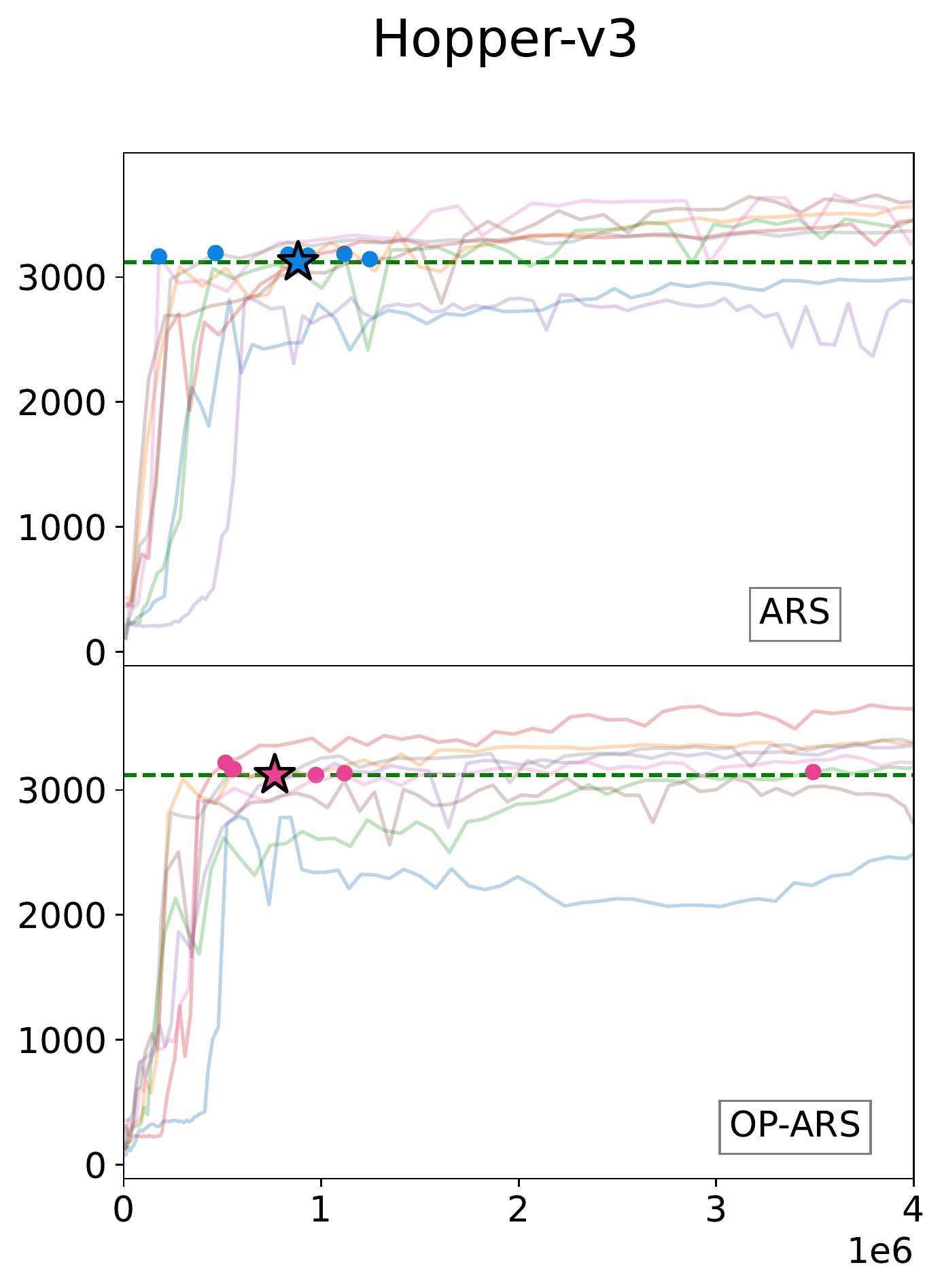}
     \end{subfigure}
     \begin{subfigure}[c]{0.25\textwidth}
         \centering
         \includegraphics[width=\textwidth]{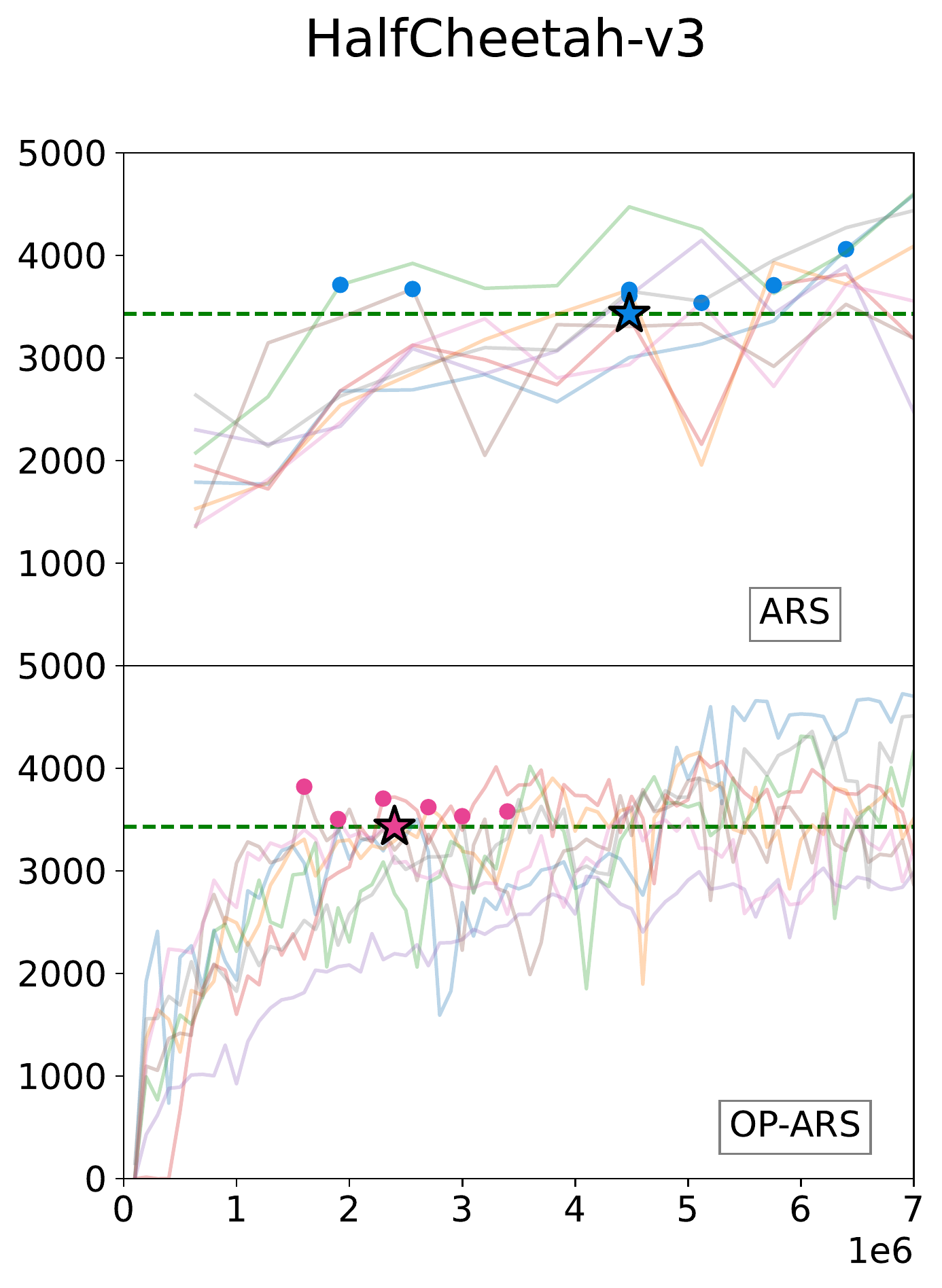}
     \end{subfigure}
     \begin{subfigure}[c]{0.26\textwidth}
         \centering
         \includegraphics[width=\textwidth]{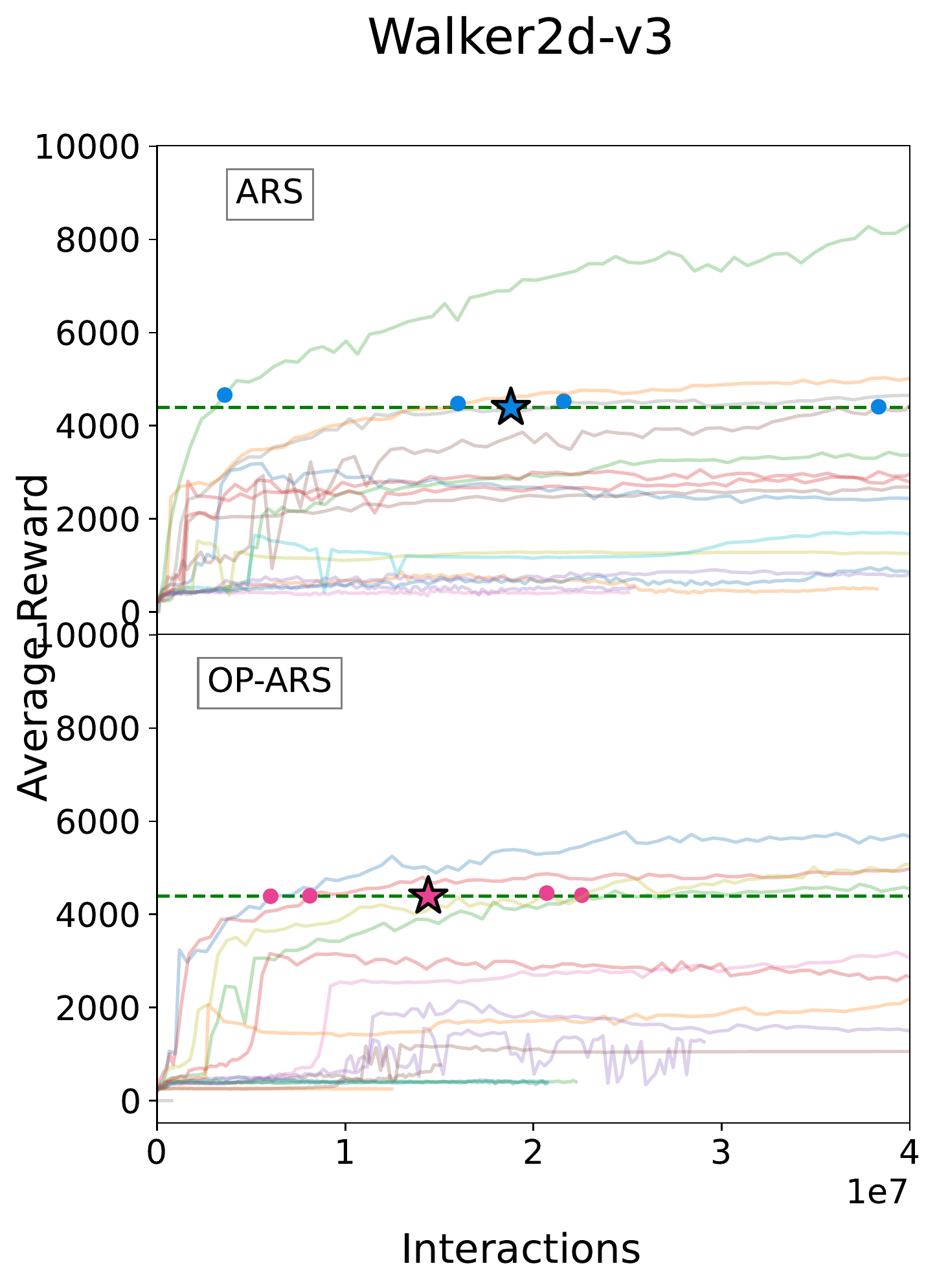}
     \end{subfigure}
     \begin{subfigure}[c]{0.25\textwidth}
         \centering
         \includegraphics[width=\textwidth]{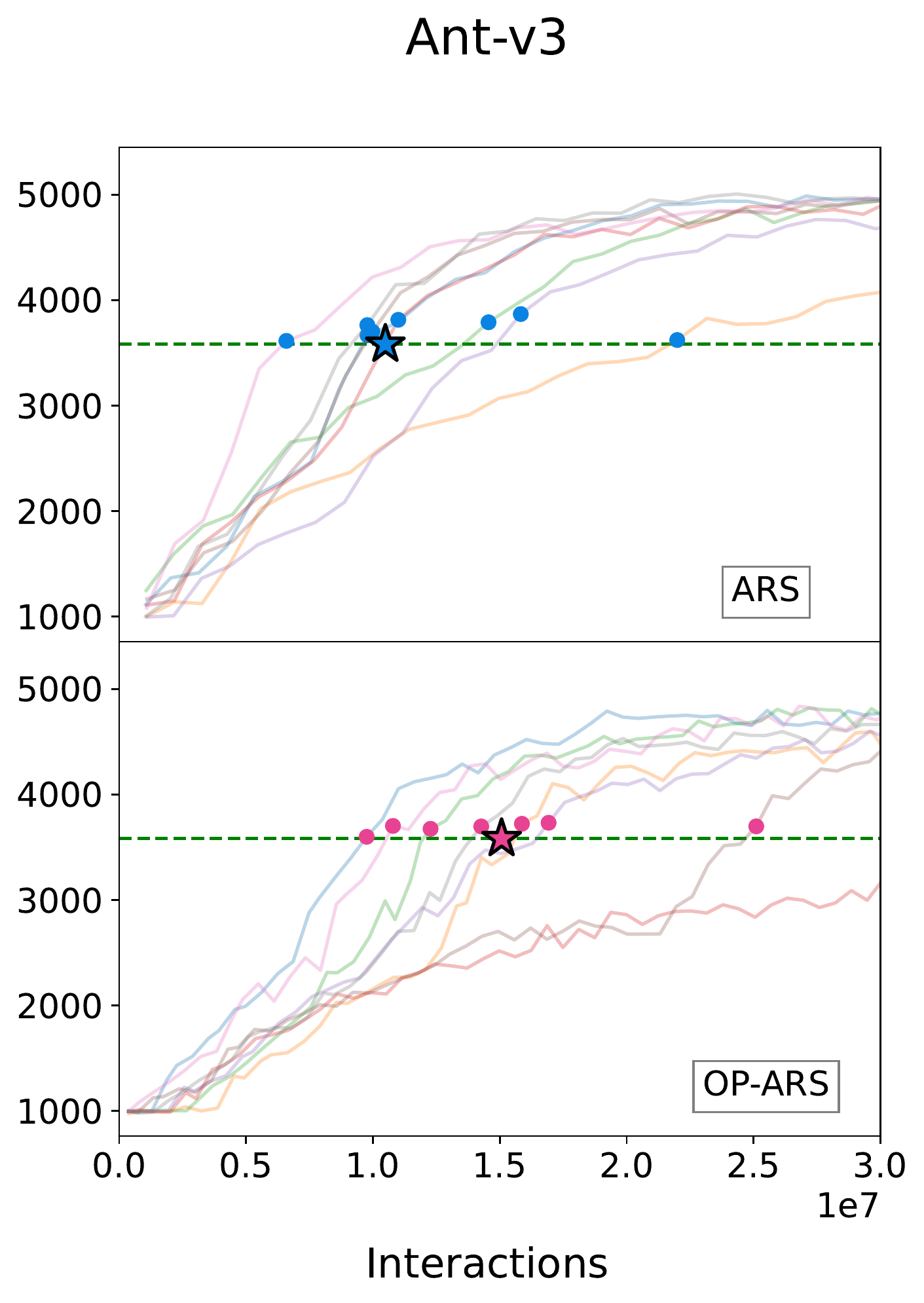}
     \end{subfigure}
     \begin{subfigure}[c]{0.25\textwidth}
         \centering
         \includegraphics[width=\textwidth]{all_seeds_ars/Humanoid-v3_split.pdf}
     \end{subfigure}
        \caption{Figures representing the trajectories of various runs of ARS and OP-ARS algorithms where the number of interactions with the environment is plotted against the average reward.}
        \label{fig:appendix.all_seeds_ars}
\end{figure*}

As the box-plots depict the median and other quantiles of timesteps required to reach the threshold, the overall progress is not very clear, hence we plot Figure \ref{fig:appendix.all_seeds_tres} and Figure \ref{fig:appendix.all_seeds_ars} to illustrate the same. Again, it can be seen clearly that, the proposed methods reach the reward threshold much faster than the other methods except for Ant in OP-ARS.

\newpage

\begin{figure*}[h!]
     \centering
     \begin{subfigure}[c]{0.27\textwidth}
         \centering
         \includegraphics[width=\textwidth]{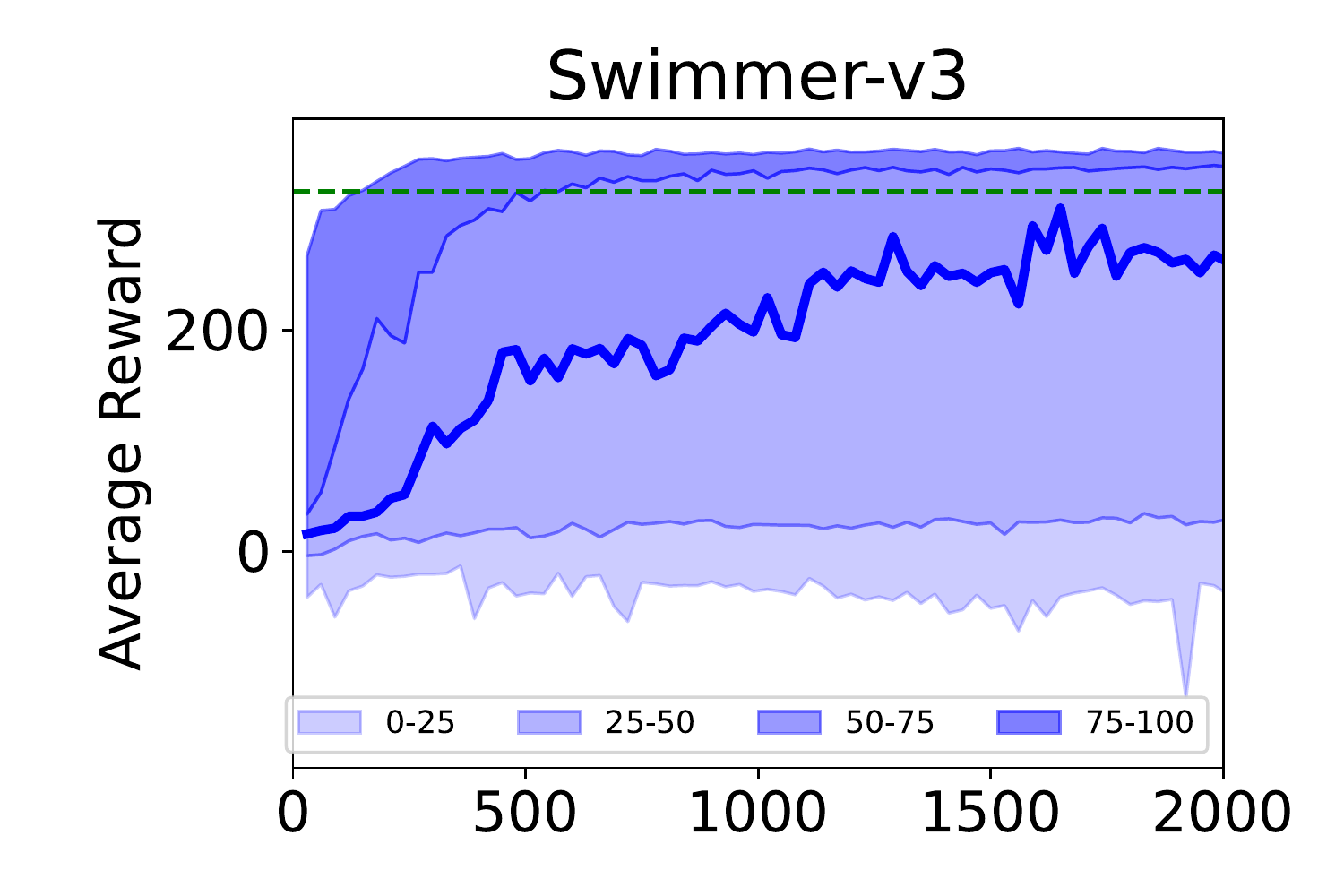}
     \end{subfigure}
     \begin{subfigure}[c]{0.27\textwidth}
         \centering
         \includegraphics[width=\textwidth]{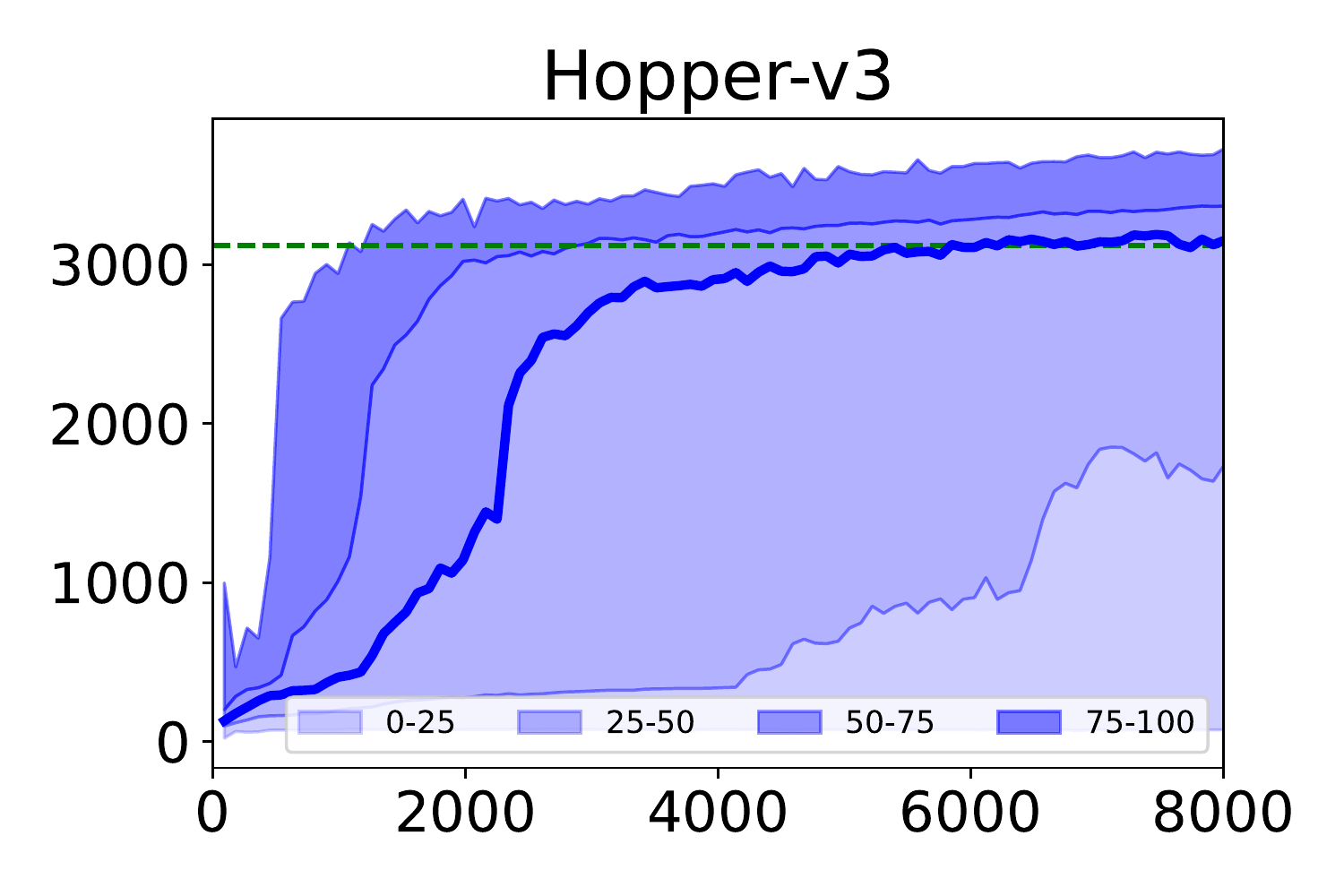}
     \end{subfigure}
     \begin{subfigure}[c]{0.27\textwidth}
         \centering
         \includegraphics[width=\textwidth]{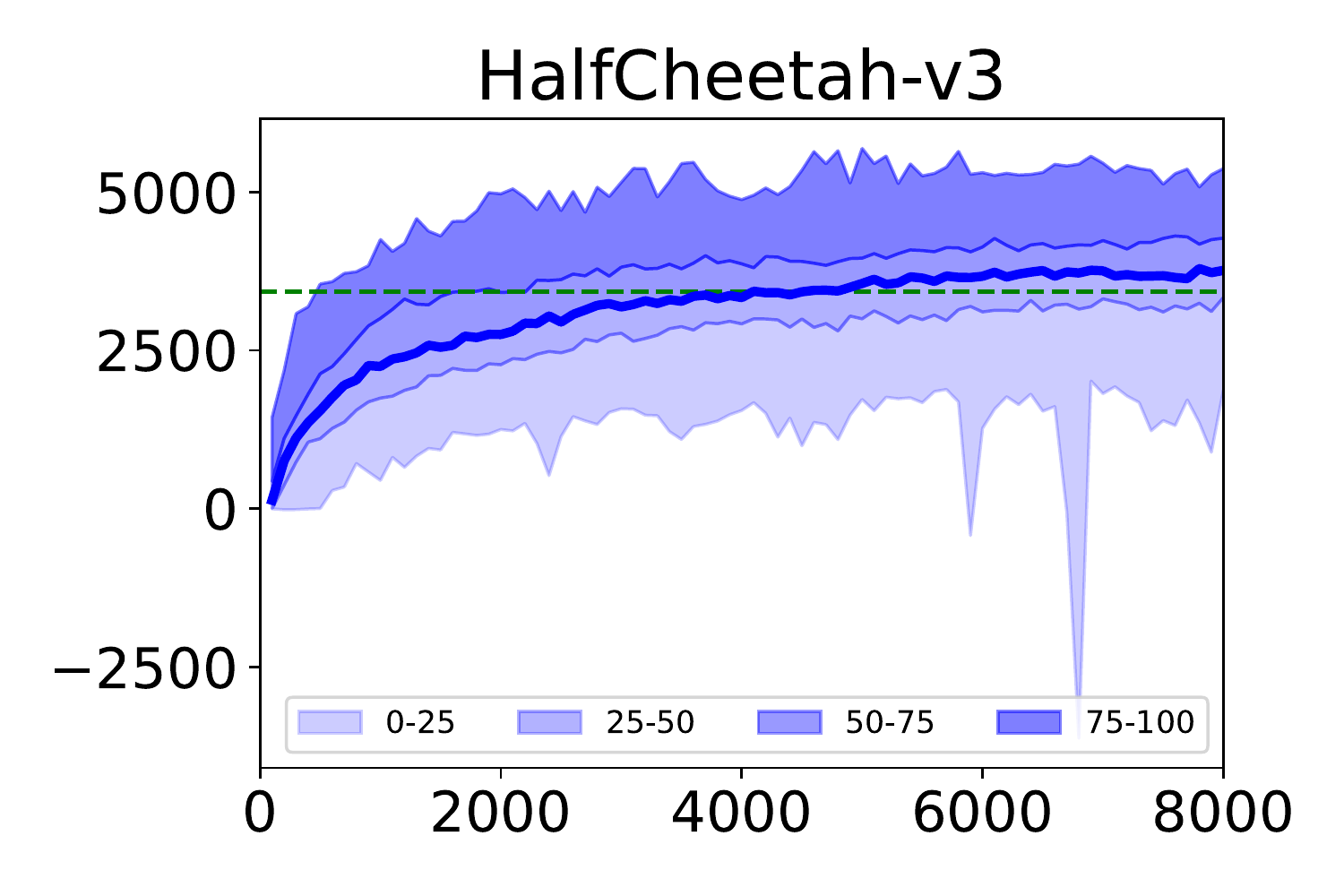}
     \end{subfigure}
     \begin{subfigure}[c]{0.27\textwidth}
         \centering
         \includegraphics[width=\textwidth]{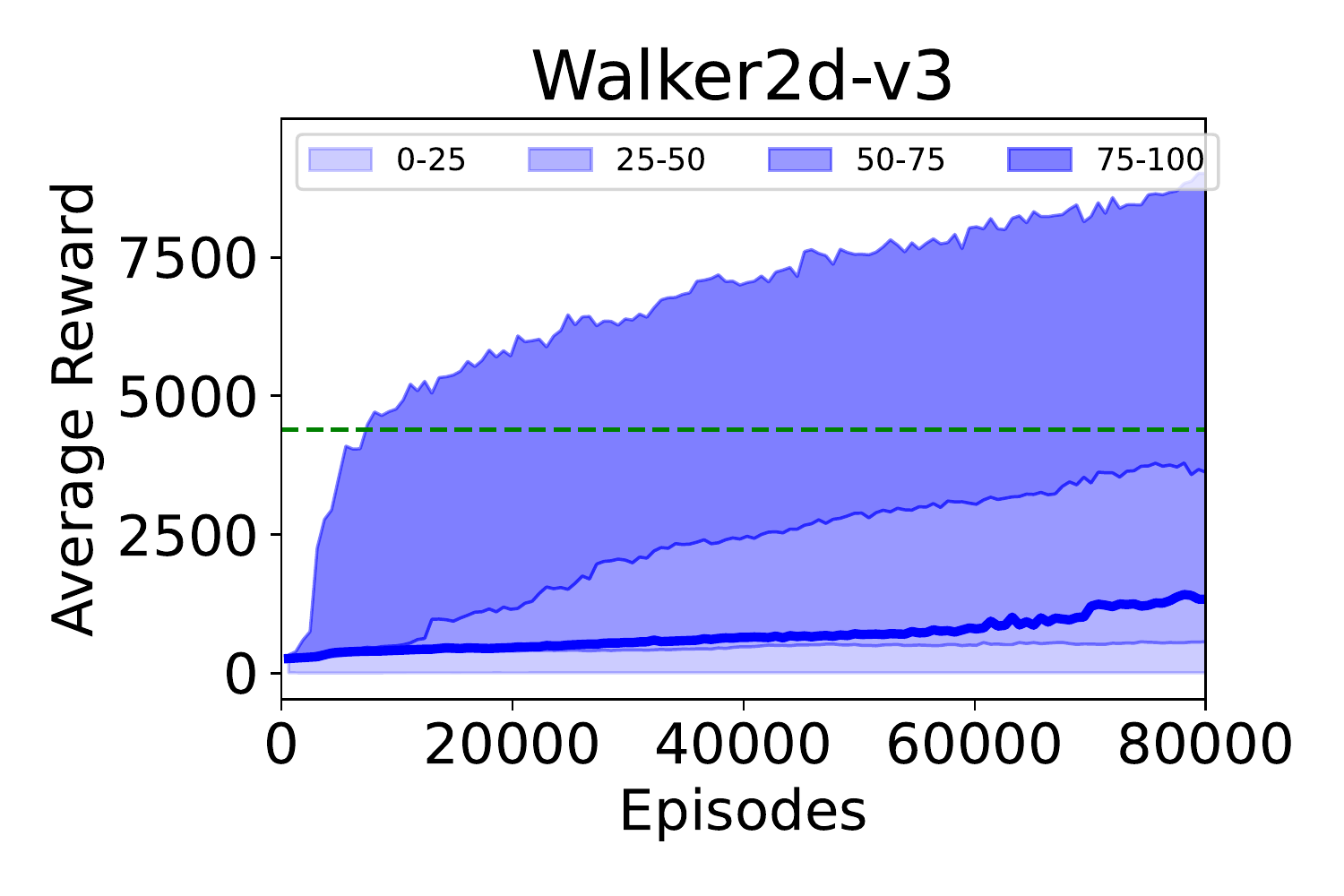}
     \end{subfigure}
     \begin{subfigure}[c]{0.27\textwidth}
         \centering
         \includegraphics[width=\textwidth]{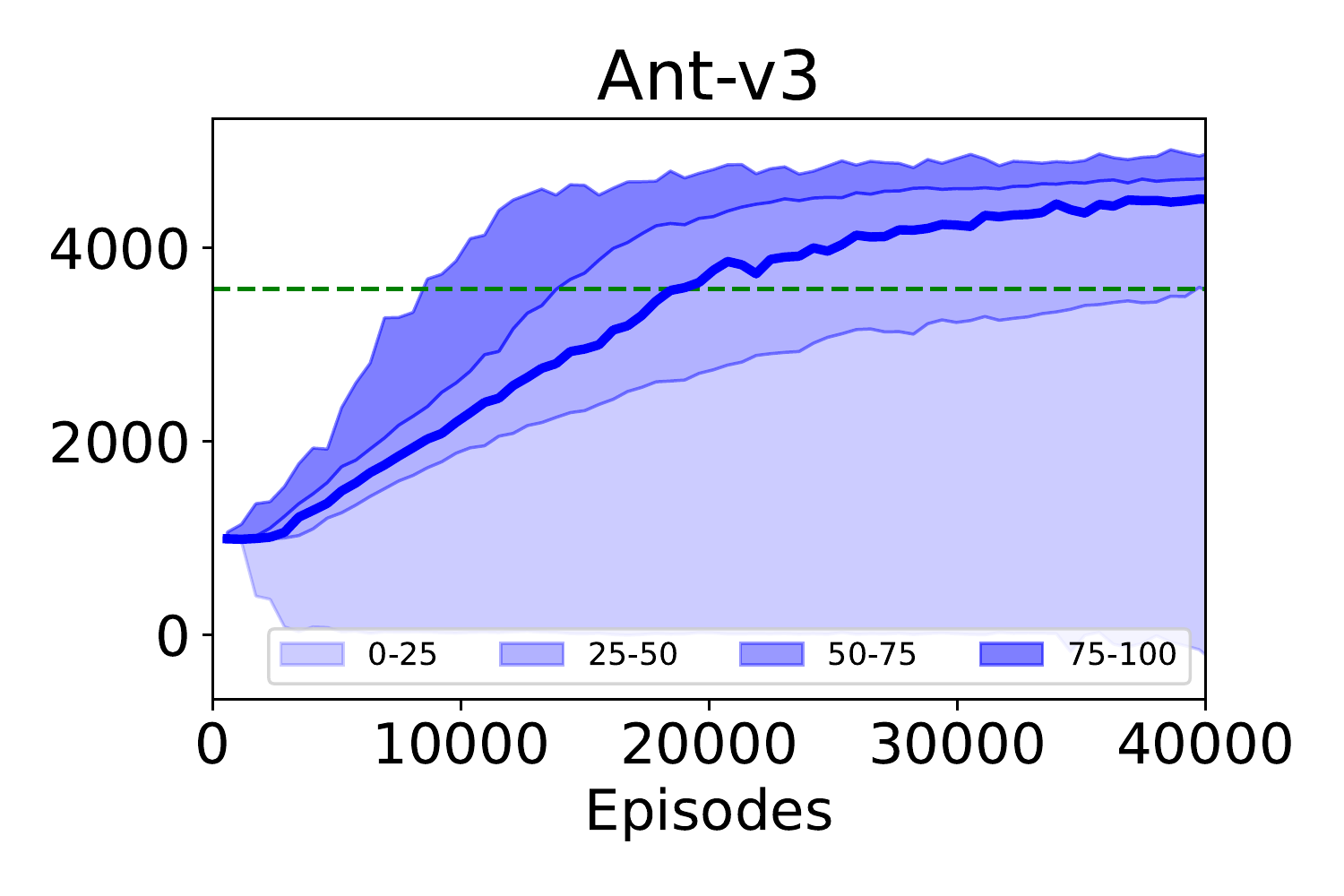}
     \end{subfigure}
     \begin{subfigure}[c]{0.27\textwidth}
         \centering
         \includegraphics[width=\textwidth]{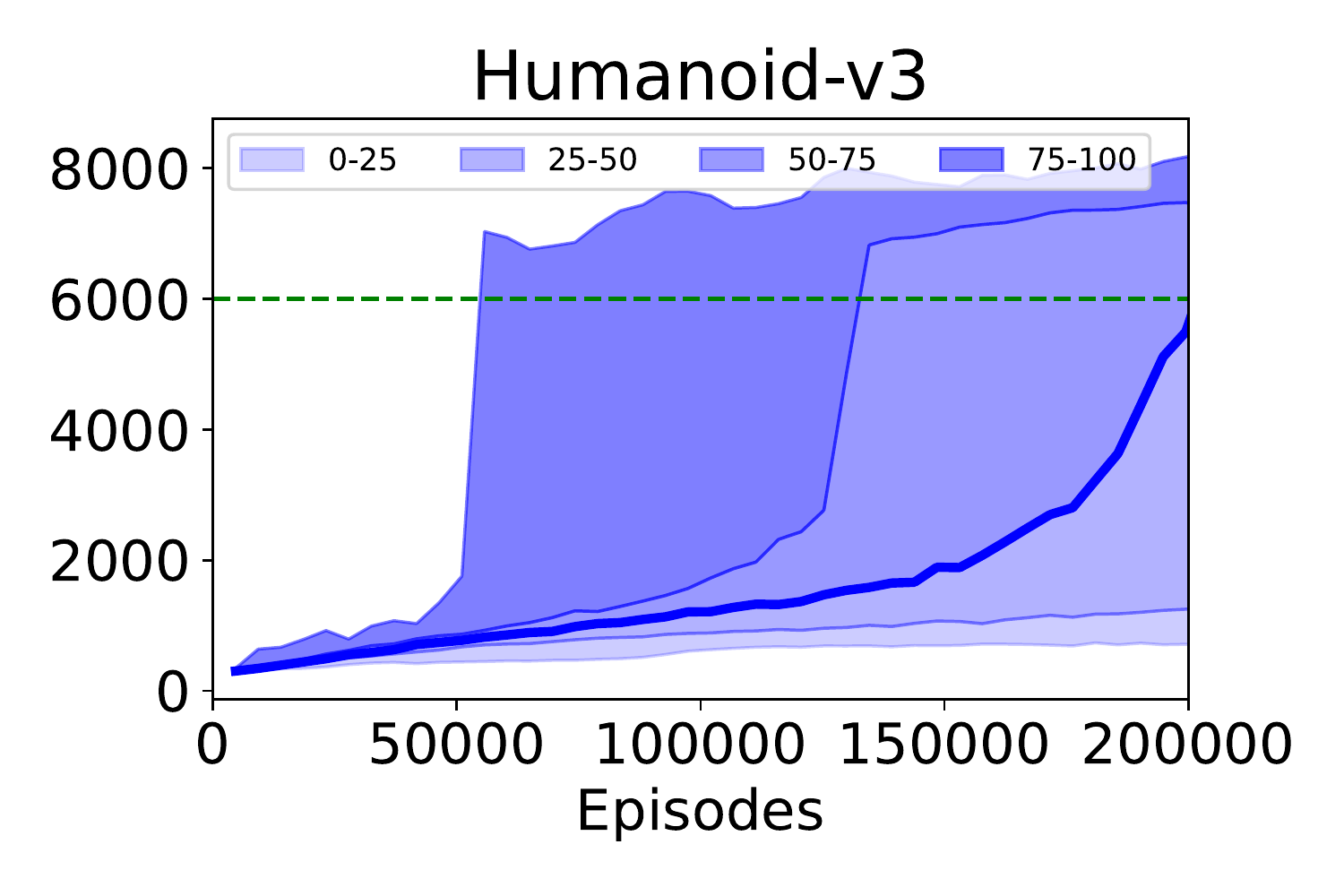}
     \end{subfigure}
        \caption{Evaluation over 100 seeds of OP-ARS. Average reward is plotted against Episodes. The thick blue line represents the median curve, and the shaded region corresponds to the percentiles mentioned in the legend.}
        \label{fig:100_seeds_ars}
\end{figure*}

\begin{figure*}[h!]
     \centering
     \begin{subfigure}[c]{0.27\textwidth}
         \centering
         \includegraphics[width=\textwidth]{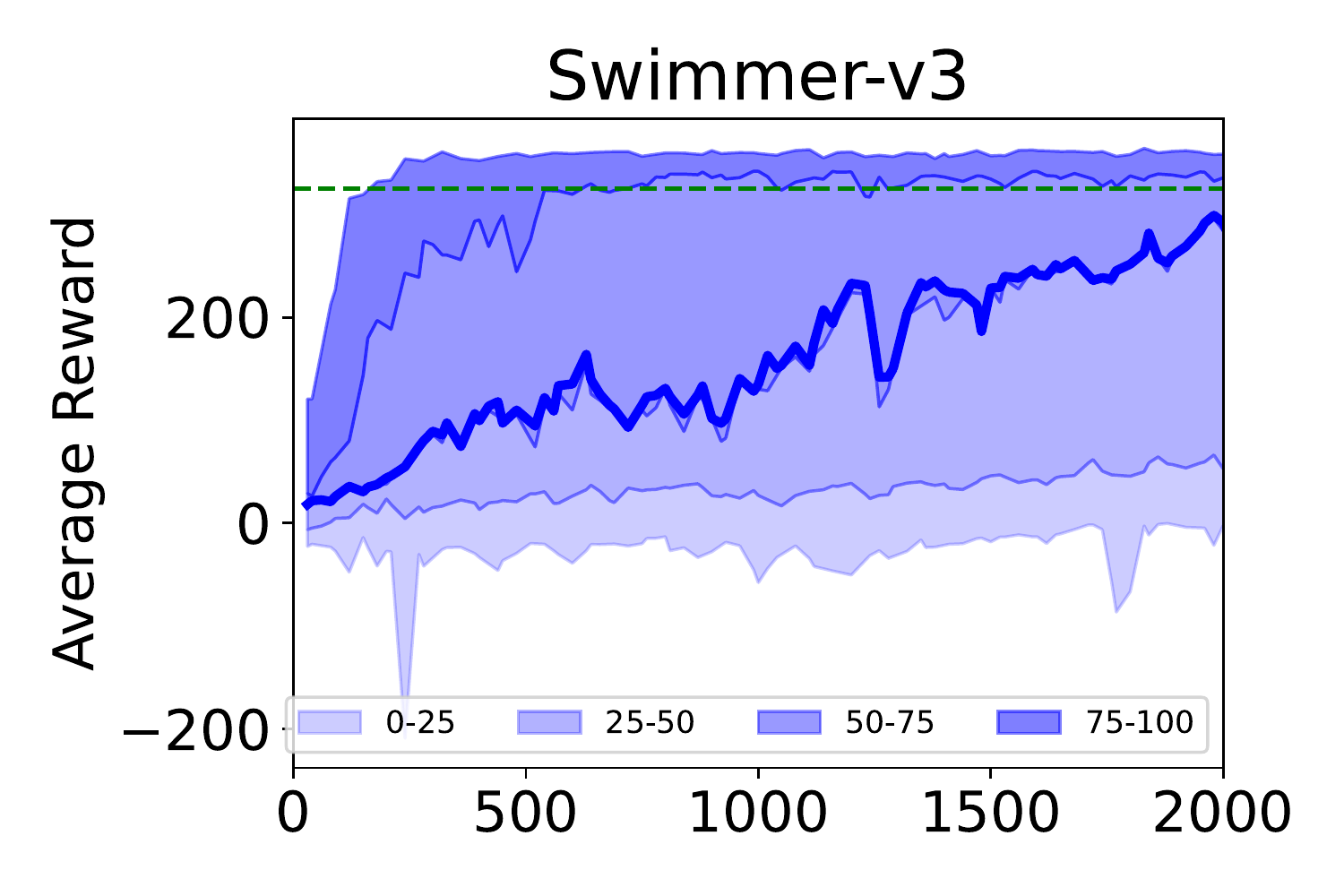}
     \end{subfigure}
     \begin{subfigure}[c]{0.27\textwidth}
         \centering
         \includegraphics[width=\textwidth]{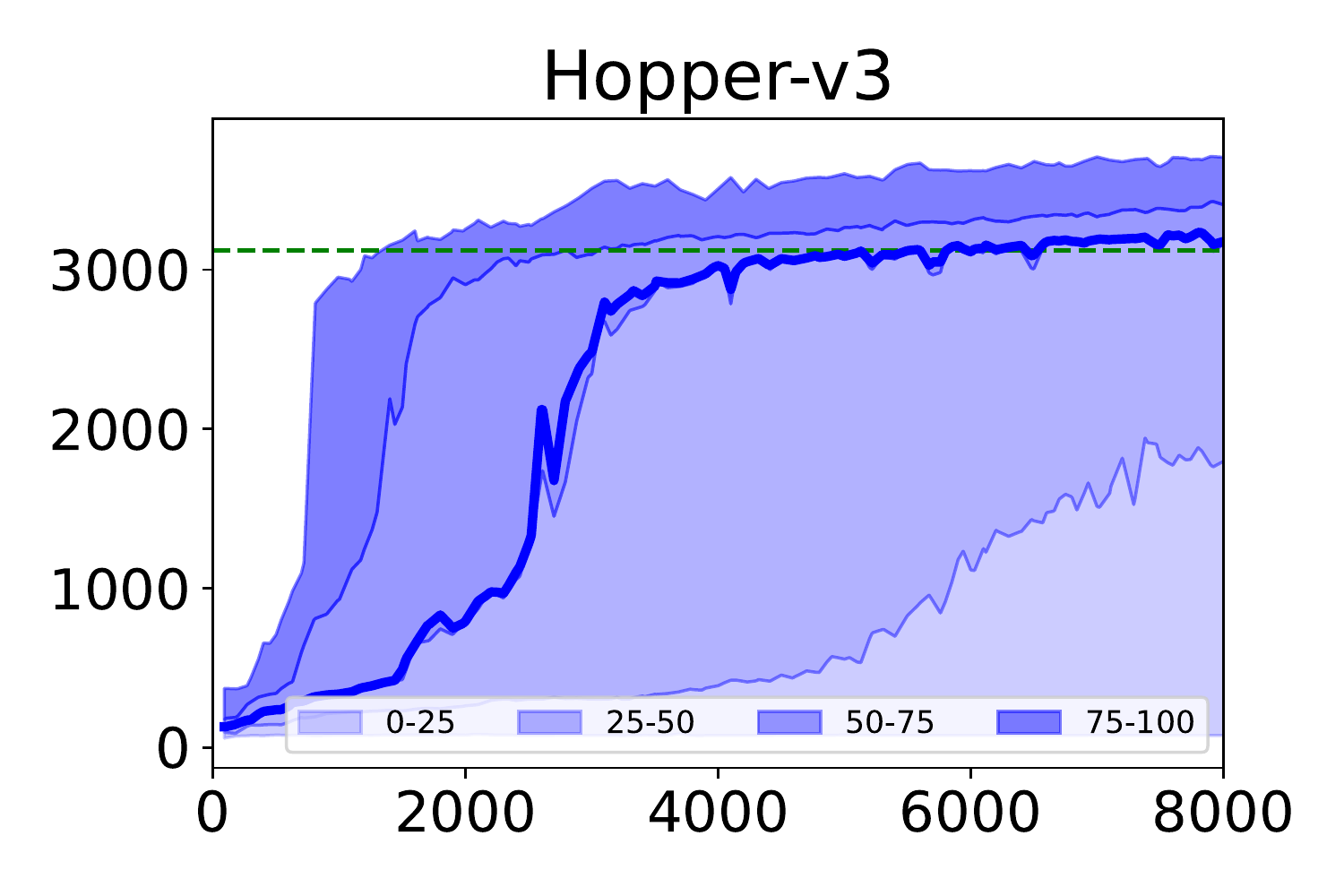}
     \end{subfigure}
     \begin{subfigure}[c]{0.27\textwidth}
         \centering
         \includegraphics[width=\textwidth]{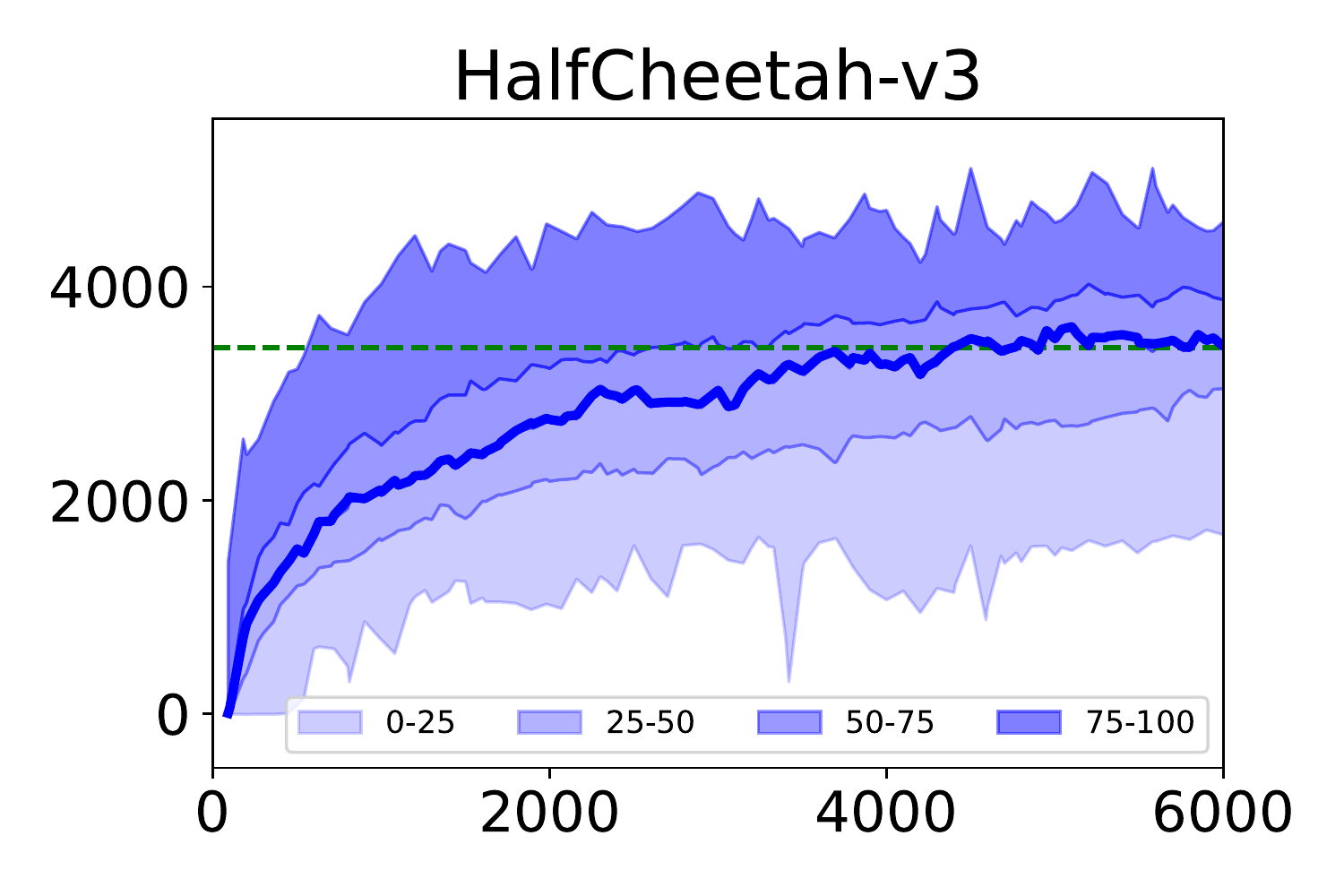}
     \end{subfigure}
     \begin{subfigure}[c]{0.27\textwidth}
         \centering
         \includegraphics[width=\textwidth]{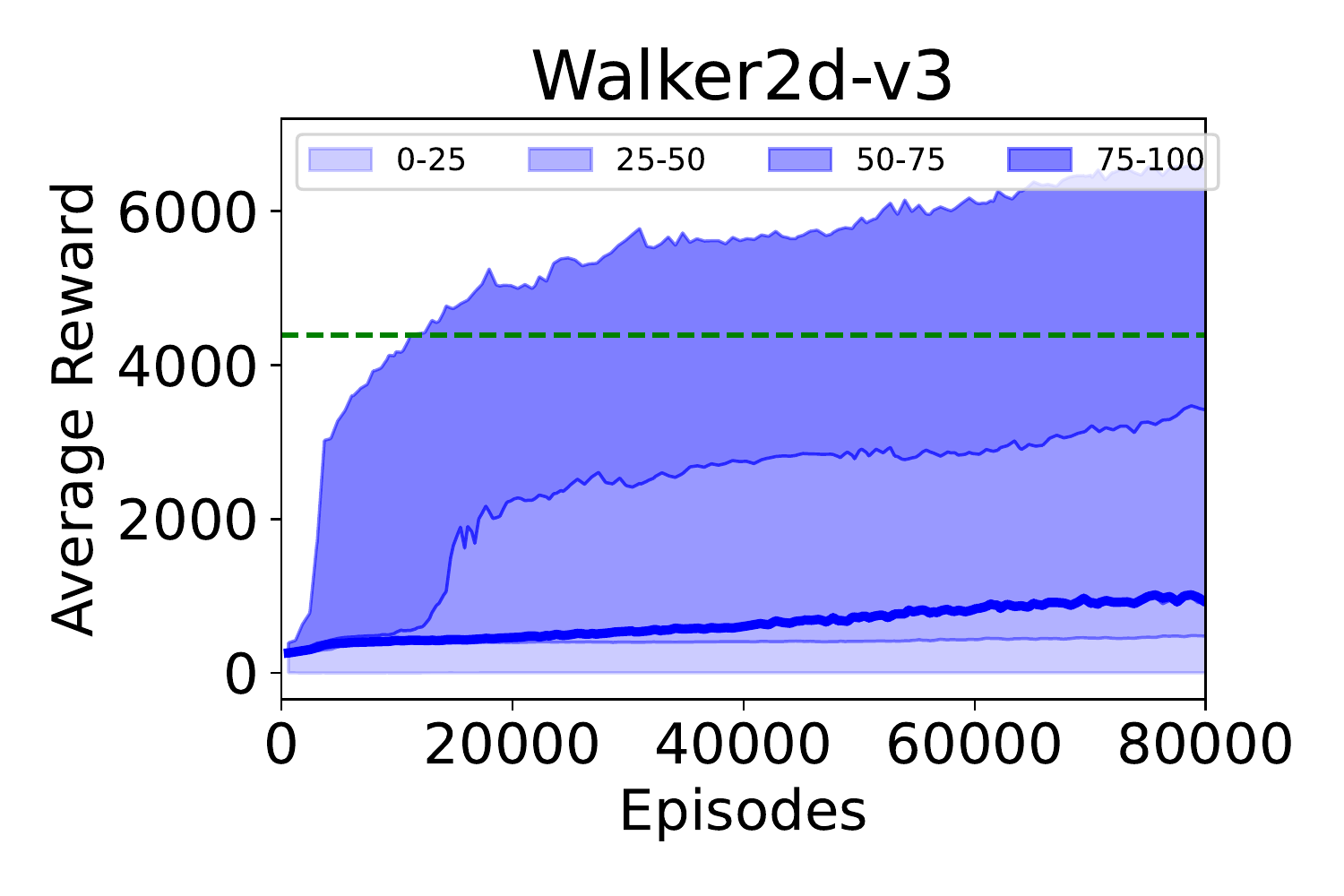}
     \end{subfigure}
     \begin{subfigure}[c]{0.27\textwidth}
         \centering
         \includegraphics[width=\textwidth]{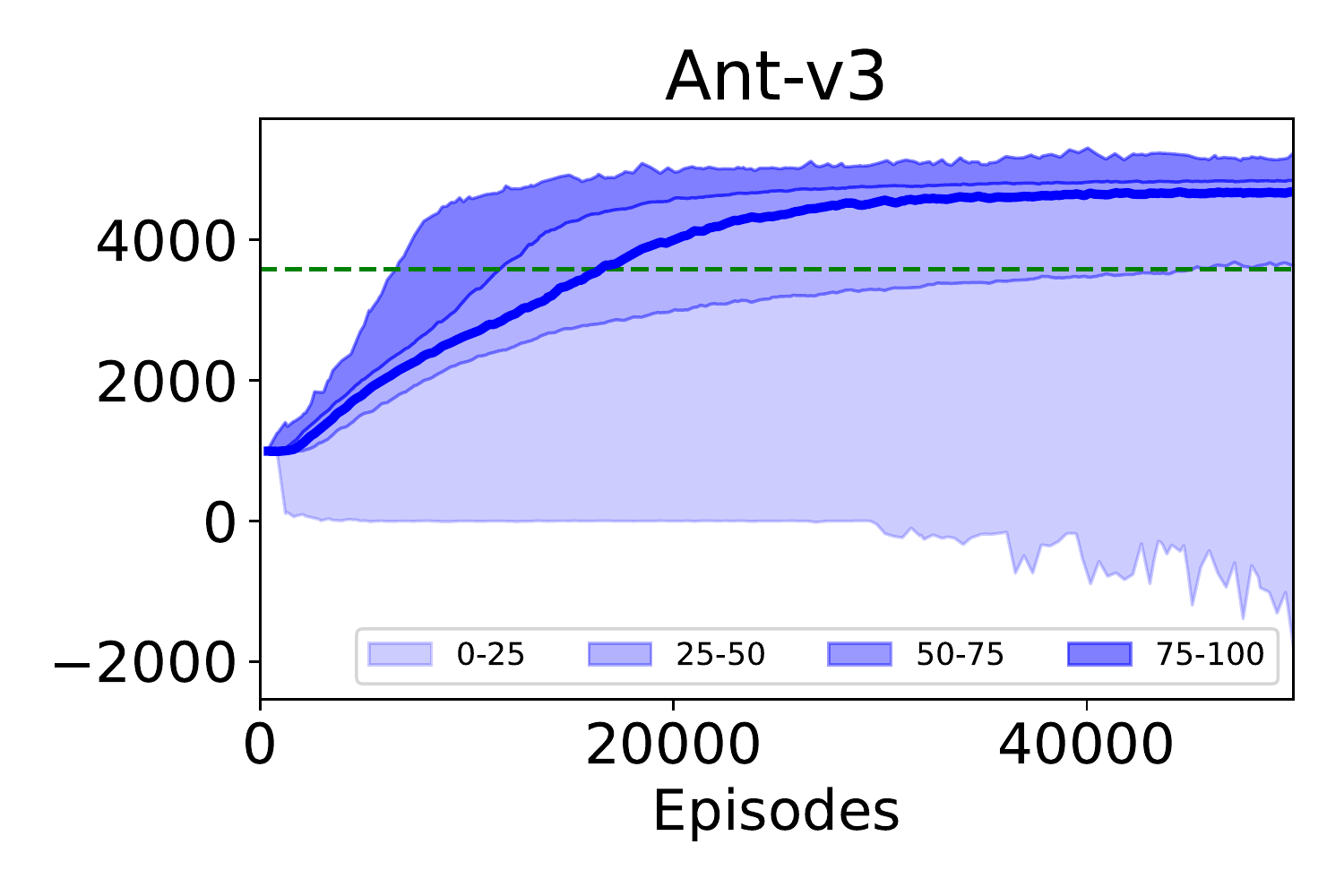}
     \end{subfigure}
     \begin{subfigure}[c]{0.27\textwidth}
         \centering
         \includegraphics[width=\textwidth]{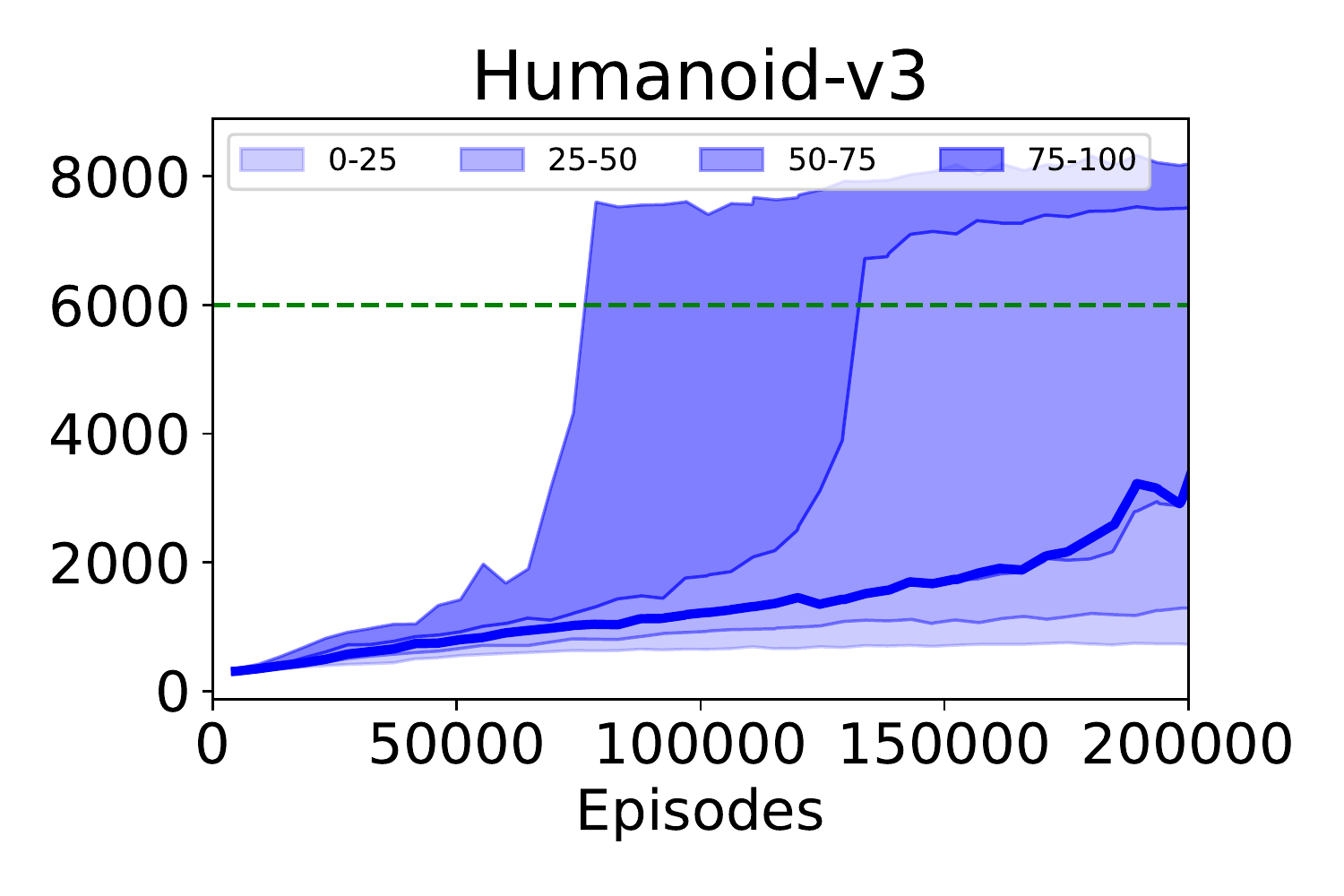}
     \end{subfigure}
        \caption{Evaluation of sensitivity to hyperparameters of OP-ARS. Average reward is plotted against Episodes. The thick blue curve is the median, and the shaded region is of percentiles given in the legend.}
        \label{fig:hyperparam_sensitivity_ars}
\end{figure*}

Next, as discussed in Section \ref{subsec:random_seeds_and_hyper_choice}, we perform experiments to verify robustness of OP-ARS to seeds and hyper-parameter choices. We had presented plots of only Swimmer, Hopper and HalfCheetah in the main paper. Here, the Figure \ref{fig:100_seeds_ars} represents the robustness of OP-ARS in all the environments except Walker. This is due to low success rates in Walker environment. Figure \ref{fig:hyperparam_sensitivity_ars} represent the plots resulting from various combination of hyper-parameters shown in Table \ref{tab:all_hypers}. These plots match the ones with plots in Figure \ref{fig:100_seeds_ars} confirming that the algorithm’s performance is not sensitive to hyperparameter choices.

\clearpage

\bibliographystyle{named}
\bibliography{mypaper}

\end{document}